\documentclass{article}

\usepackage{etoolbox}
\newcommand{\arxiv}[1]{\iftoggle{icml}{}{#1}}
\newcommand{\icml}[1]{\iftoggle{icml}{#1}{}}
\newtoggle{icml}
\global\toggletrue{icml}
\global\togglefalse{icml}

\icml{
\PassOptionsToPackage{dvipsnames}{xcolor} 
\usepackage[accepted]{icml2024}
}

\newcommand{\loose}{\looseness=-1}

\usepackage[utf8]{inputenc} 
\usepackage[T1]{fontenc}    
\usepackage{url}            
\usepackage{booktabs}       
\usepackage{amsfonts}       
\usepackage{nicefrac}       
\usepackage{microtype}      

\usepackage{tocloft}            

\usepackage{enumitem}
\setlist[itemize]{}

\usepackage{breakcites}

\newtoggle{draft}
\togglefalse{draft}

\usepackage{mathrsfs}

\usepackage{algorithm}
\usepackage{verbatim}
\usepackage[noend]{algpseudocode}

\usepackage{multicol}

\usepackage{colortbl}

\usepackage{setspace}

\usepackage{transparent}

\usepackage{inconsolata}
\usepackage[scaled=.90]{helvet}
\usepackage{xspace}

\usepackage{pifont}
\arxiv{
\usepackage[letterpaper, left=1in, right=1in, top=1in, bottom=1in]{geometry}
\PassOptionsToPackage{hypertexnames=false}{hyperref}  
\usepackage{parskip}
\usepackage[dvipsnames]{xcolor}
\usepackage[colorlinks]{hyperref}
\hypersetup{
    citecolor=[RGB]{50,100,170},
    linkcolor=[RGB]{50,100,170},
    urlcolor=[RGB]{255,102,178}}
}

\icml{
\usepackage[colorlinks]{hyperref}
\hypersetup{
    citecolor=[RGB]{50,100,170},
    linkcolor=[RGB]{50,100,170},
    urlcolor=[RGB]{255,102,178}}

}

\usepackage{microtype}
\usepackage{hhline}

\makeatletter
\newcommand{\neutralize}[1]{\expandafter\let\csname c@#1\endcsname\count@}
\makeatother

\usepackage{algorithm}

\arxiv{
\usepackage{natbib}
\bibliographystyle{plainnat}
\bibpunct{(}{)}{;}{a}{,}{,}
}

\usepackage{amsthm}
\usepackage{mathtools}
\usepackage{amsmath}
\usepackage{bbm}
\usepackage{amsfonts}
\usepackage{amssymb}

\usepackage{xpatch}

\usepackage{thmtools}
\usepackage{thm-restate}
\declaretheorem[name=Theorem,parent=section]{theorem}
\declaretheorem[name=Lemma,parent=section]{lemma}
\declaretheorem[name=Assumption, parent=section]{assumption}
\declaretheorem[name=Condition, parent=section]{condition}

\declaretheorem[name=Remark,style=definition, parent=section]{remark}
\declaretheorem[name=Proposition, parent=section]{proposition}

\makeatletter
  \renewenvironment{proof}[1][Proof]%
  {%
   \par\noindent{\bfseries\upshape {#1.}\ }%
  }%
  {\qed\newline}
  \makeatother

\theoremstyle{definition}  

\newtheorem{corollary}{Corollary}[section]

\theoremstyle{plain}
\newtheorem{definition}{Definition}[section]

\xpatchcmd{\proof}{\itshape}{\normalfont\proofnameformat}{}{}
\newcommand{\proofnameformat}{\bfseries}

\usepackage[nameinlink,capitalize]{cleveref}

\newcommand{\pref}[1]{\cref{#1}}

\newcommand{\savehyperref}[2]{\texorpdfstring{\hyperref[#1]{#2}}{#2}}

\renewcommand{\eqref}[1]{\texorpdfstring{\hyperref[#1]{(\ref*{#1})}}{(\ref*{#1})}}

\crefformat{equation}{#2Eq. (#1)#3}
\Crefformat{equation}{#2Eq. (#1)#3}

\Crefformat{figure}{#2Figure #1#3}
\Crefname{assumption}{Assumption}{Assumptions}
\Crefformat{assumption}{#2Assumption #1#3}
\Crefname{subsubsection}{Section}{Sections}
\crefformat{subsubsection}{#2Section #1#3}
\Crefformat{subsubsection}{#2Section #1#3}

\usepackage{crossreftools}
\usepackage{xparse}

\ExplSyntaxOn
\DeclareDocumentCommand{\XDeclarePairedDelimiter}{mm}
 {
  \__egreg_delimiter_clear_keys: 
  \keys_set:nn { egreg/delimiters } { #2 }
  \use:x 
   {
    \exp_not:n {\NewDocumentCommand{#1}{sO{}m} }
     {
      \exp_not:n { \IfBooleanTF{##1} }
       {
        \exp_not:N \egreg_paired_delimiter_expand:nnnn
         { \exp_not:V \l_egreg_delimiter_left_tl }
         { \exp_not:V \l_egreg_delimiter_right_tl }
         { \exp_not:n { ##3 } }
         { \exp_not:V \l_egreg_delimiter_subscript_tl }
       }
       {
        \exp_not:N \egreg_paired_delimiter_fixed:nnnnn 
         { \exp_not:n { ##2 } }
         { \exp_not:V \l_egreg_delimiter_left_tl }
         { \exp_not:V \l_egreg_delimiter_right_tl }
         { \exp_not:n { ##3 } }
         { \exp_not:V \l_egreg_delimiter_subscript_tl }
       }
     }
   }
 }

\keys_define:nn { egreg/delimiters }
 {
  left      .tl_set:N = \l_egreg_delimiter_left_tl,
  right     .tl_set:N = \l_egreg_delimiter_right_tl,
  subscript .tl_set:N = \l_egreg_delimiter_subscript_tl,
 }

\cs_new_protected:Npn \__egreg_delimiter_clear_keys:
 {
  \keys_set:nn { egreg/delimiters } { left=.,right=.,subscript={} }
 }

\cs_new_protected:Npn \egreg_paired_delimiter_expand:nnnn #1 #2 #3 #4
 {
  \mathopen{}
  \mathclose\c_group_begin_token
   \left#1
   #3
   \group_insert_after:N \c_group_end_token
   \right#2
   \tl_if_empty:nF {#4} { \c_math_subscript_token {#4} }
 }
\cs_new_protected:Npn \egreg_paired_delimiter_fixed:nnnnn #1 #2 #3 #4 #5
 {
  \mathopen{#1#2}#4\mathclose{#1#3}
  \tl_if_empty:nF {#5} { \c_math_subscript_token {#5} }
 }
\ExplSyntaxOff

\XDeclarePairedDelimiter{\supnorm}{
  left=\lVert,
  right=\rVert,
  subscript=\infty
  }

\usepackage{mkolar_definitions}

\newcommand{\ccov}{C_\mathsf{cov}}
\newcommand{\ccovpd}{C_\mathsf{cov}^{\mathsf{pd}}}
\newcommand{\Mreset}{\Mcal_\mathsf{reset}}
\newcommand{\pif}{\pi^\mathsf{f}}
\newcommand{\pib}{\pi^\mathsf{b}}
\newcommand{\IPM}{\mathsf{IPM}}
\newcommand{\piunif}{\pi^\mathsf{unif}}
\newcommand{\Psdp}{\textsc{Psdp}} 
 
\newcommand{\Fail}{\textsc{Fail}} 
\newcommand{\Cpi}{\textsc{Cpi}} 
\newcommand{\rlpd}{\textsc{Rlpd}} 

\newcommand{\Foobar}{\textsc{Foobar}} 
\newcommand{\HyQ}{\textsc{HyQ}}

\newcommand{\epsforward}{\varepsilon^\mathsf{for}} 
\newcommand{\epsback}{\varepsilon^\mathsf{back}} 
\newcommand{\hyrlo}{\textsc{HyRLO}}
\newcommand{\ilfo}{\textsc{ILfO}}
\newcommand{\TV}{\mathsf{TV}}
\newcommand{\mathand}{\quad\text{and}\quad}
\newcommand{\picomp}{\pi^\mathsf{comp}}

\newcommand{\muit}{\mu^\mathsf{it}}
\newcommand{\Don}{\Dcal^\mathsf{on}}
\newcommand{\Doff}{\Dcal^\mathsf{off}}

\newcommand*\colourcheck[1]{%
  \expandafter\newcommand\csname #1check\endcsname{\textcolor{#1}{\ding{52}}}%
}
\colourcheck{yellow}
\colourcheck{green}
\colourcheck{red}

\newcommand*\colourquestion[1]{%
  \expandafter\newcommand\csname #1question\endcsname{\textcolor{#1}{\textbf{?}}}%
}

\colourquestion{red}

\usepackage{makecell}

\usepackage[suppress]{color-edits}
 \addauthor{jab}{ForestGreen}
 \addauthor{ys}{TealBlue}
 \addauthor{as}{BurntOrange}

 \arxiv{
\usepackage[final]{showlabels}

}

\makeatletter
\let\OldStatex\Statex
\renewcommand{\Statex}[1][3]{%
  \setlength\@tempdima{\algorithmicindent}%
  \OldStatex\hskip\dimexpr#1\@tempdima\relax}
\makeatother

\usepackage{accents}
\usepackage{wrapfig}
\usepackage{tikz}
\usetikzlibrary{decorations.pathreplacing}

 \addtocontents{toc}{\protect\setcounter{tocdepth}{0}}

\let\oldparagraph\paragraph
\renewcommand{\paragraph}[1]{\oldparagraph{#1.}}

\algrenewcommand\algorithmicrequire{\textbf{require}}
\newcommand{\algcommentlight}[1]{\textcolor{blue!70!black}{\transparent{0.5}\small{\texttt{\textbf{//\hspace{2pt}#1}}}}}

\usepackage{yfonts}

\arxiv{
\title{Hybrid Reinforcement Learning from Offline Observation Alone}
\author{Yuda Song \\
{\small Carnegie Mellon University}\\
{\small \texttt{yudas@cs.cmu.edu}}
\and
J. Andrew Bagnell \\
{\small Aurora Innovation, Carnegie Mellon University}\\
{\small \texttt{dbagnell@cs.cmu.edu}}
\and 
Aarti Singh\\
{\small Carnegie Mellon University}\\
{\small \texttt{aarti@cs.cmu.edu}}
}
\date{}
}

\icml{
\icmltitlerunning{Hybrid Reinforcement Learning from Offline Observation Alone}
}

\begin{document}

\arxiv{\maketitle}

\icml{
\twocolumn[
\icmltitle{Hybrid Reinforcement Learning from Offline Observation Alone}


\icmlsetsymbol{equal}{*}

\begin{icmlauthorlist}
\icmlauthor{Yuda Song}{cmu}
\icmlauthor{J. Andrew Bagnell}{cmu,aurora}
\icmlauthor{Aarti Singh}{cmu}
\end{icmlauthorlist}

\icmlaffiliation{cmu}{Carnegie Mellon University}
\icmlaffiliation{aurora}{Aurora Innovation}

\icmlcorrespondingauthor{Yuda Song}{yudas@cs.cmu.edu}


\vskip 0.3in
]
\printAffiliationsAndNotice{}  
}

\begin{abstract}

We consider the hybrid reinforcement learning setting where the agent has access to both offline data and online interactive access. While Reinforcement Learning (RL) research typically assumes offline data contains complete action, reward and transition information, datasets with only state information (also known as \textit{observation-only} datasets) are more general, abundant and practical. This motivates our study of the \textit{hybrid RL with observation-only offline dataset} framework. 
While the task of competing with the best policy ``covered'' by the offline data can be solved if a \textit{reset} model of the environment is provided (i.e., one that can be reset to any state), we show evidence of hardness when only given the weaker \textit{trace} model (i.e., one can only reset to the initial states and must produce full traces through the environment), without further assumption of \textit{admissibility} of the offline data. Under the admissibility assumptions-- that the offline data could actually be produced by the policy class we consider-- we propose the first algorithm in the trace model setting that provably matches the performance of algorithms that leverage a reset model. We also perform proof-of-concept experiments that suggest the effectiveness of our algorithm in practice. \loose

\end{abstract}

\section{Introduction}
\label{sec:intro}
Recently, explosive growth in the availability of offline data for interactive decision making problems \citep{dasari2019robonet,qin2022neorl,mathieu2023alphastar,padalkar2023open}, combined with an ability to interact with the environment for feedback, 
led to the advancement of hybrid Reinforcement Learning (hybrid RL)  \citep{ross2012agnostic,song2022hybrid}. This setup blends the exploratory strengths of offline data with the ability to adapt the data distribution using online interaction with the environment. 
Previously, RL research has tended to focus on either purely offline or online regimes, each with its inherent challenges.   Offline learning, while benefiting from the exploration and potential expert ``advice'' implicit in a large offline dataset, often suffers from instability due to distribution shifts \citep{wang2021instabilities}. Online RL benefits from feedback from the environment but at the cost of increased complexity, both statistically and computationally \citep{du2019good, kane2022computational}, due to the requirement of global exploration. Hybrid RL benefits from the synergy of combining both data sources. Earlier studies in this domain have predominantly utilized canonical offline datasets \citep{levine2020offline}, with extensive information of state, action, reward, and subsequent state. This comprehensive data approach has proven beneficial, highlighting the statistical and computational superiority of hybrid RL \citep{song2022hybrid,hu2023reboot}, and its robustness against distribution shift \citep{wagenmaker2023leveraging,ren2023hybrid}.

However, the necessity for expansive datasets in such a rich format poses a significant barrier. In practice, most large-scale datasets exist in the format of videos \citep{grauman2022ego4d,grauman2023ego} (imagine using video demonstrations from Youtube). The requirement of annotated actions in the dataset is restrictive because actions do not generalize between different modalities: one should not expect to naively apply human actions to robot actuators, and different robots have different action spaces as well. The approach of collecting different actions for different modalities with human labors \citep{padalkar2023open} is challenging to scale as the observation-only datasets. \loose

\begin{table*}[t]
        \caption{Comparisons of hybrid RL frameworks and algorithms. We compare the sample complexity, problem setting and assumptions required by each work. Our setting assumes weaker requirement on the offline data and model access (defined in \pref{sec:prelim}), while requiring the admissibility assumption (\pref{assum:admissibility}) on the offline data. We show that without the admissibility assumption, the problem might exhibit exponential sample complexity separation between the trace model and reset model in \pref{prop:hard_data} and \pref{prop:hard_tv}. Previous hybrid RL with trace model analysis \citep{song2022hybrid} requires \emph{explicit structural assumptions} on the MDP and value function \citep{du2021bilinear} (characterized as the additional $d$ parameter in their sample complexity result), while our analysis does not require such assumptions. Finally, we consider value-based learning, so $Q^\ast$ realizability denotes the optimal Q-function is contained in the function class, while $Q^\pi$ realizability requires the function class contains the Q-function for all deterministic policies (for example, all $A^S$ many policies in tabular MDPs). Finally, we note that all methods only require
        single policy coverage. The original \Psdp{} paper \citep{bagnell2003policy} did not perform the analysis under single policy coverage, and the first \Psdp{} with single policy coverage analysis can be found in \citet{bagnell2004learning,scherrer2014approximate}.}
        \centering
        \arxiv{\begin{tabular}{p{2.2cm}ccccccc}}
        \icml{\begin{tabular}{p{3.7cm}ccccccc}}
        \toprule
        & \makecell{Sample\\ Complexity}& \makecell{Offline\\ Data} & Admissibility &Model &  \makecell{Bellman\\ Complete} & \makecell{Realizability}\\
        \midrule
        \HyQ~\citep{song2022hybrid} & $\frac{C_{\mathsf{be}}^2H^5A{\color{red}d}\log(|\Fcal|/\delta)}{\epsilon^2}$ &Canonical & No &Trace  & Yes & $Q^\ast$\\
        \Psdp~\citep{bagnell2003policy} & $\frac{C_{\mathsf{pd}}^2H^5A\log(|\Fcal|/\delta)}{\epsilon^2}$ &State-only &No &Reset  & No & $Q^\pi$\\
        This work & $\frac{C_{\mathsf{pd}}^2H^5A\log(|\Fcal||\Pi|/\delta)}{\epsilon^2}$ &State-only & \makecell{Yes {\small (Hardness} \\ {\small examples if No)}}& Trace  & Yes & $Q^\pi$\\
        \bottomrule
    \end{tabular}
    \label{tab:comparison}
\end{table*}

This motivates a more general setting with a weaker offline data requirement without losing the statistical and computational benefit of hybrid RL. In this work, we initiate the study of Hybrid RL from (Offline) Observation Alone (\hyrlo) framework where the offline data only contains state\footnote{We will use the terms observation and state interchangeably.} information.
Previous studies that fall into the \hyrlo{} framework can be generally characterized in two ways: the first leverages the offline state data to perform representation learning (i.e., requires a separate pretraining stage) \citep{nair2018visual,ma2022vip,ghosh2023reinforcement}, and then to use the learned feature map to speed up the downstream online RL training.  However, we show that the state-only offline dataset, although less informative than the canonical offline data, still provides a rich signal for decision-making and not only representation learning. The second previous approach relies on a \textit{reset model} \citep{ kakade2002approximately,bagnell2003policy}, which only holds true if a simulator is available and thus does not address many common real-world scenarios. In this work, we show that we can solve \hyrlo{} \textit{without} reset model access-- i.e., with a trace model that only allows resets to the initial state.
Our approach, in contrast with earlier methods, requires a notion of \textit{admissibility} \citep{chen2019information} of the offline data which formalizes the idea that the offline data should have been generated by \textit{some} policy or mixture of policies. Indeed, \hyrlo{} fills in a missing piece theoretically where there is neither complete data, as in between canonical hybrid RL, and access only to a trace model, a much weaker and more realistic access model for RL problems. We provide a comparison overview in \pref{tab:comparison}. \loose

\arxiv{\subsection{Contributions}}
\icml{\paragraph{Contributions}}
We initiate a theoretical study of  \hyrlo{} framework and provide the first provable algorithm for \hyrlo{}. Specifically, we introduce: 
\begin{itemize}
    \arxiv{\item \textbf{Mathematical formulation.} We provide the first theoretical formulation of the \hyrlo{} in the most general terms, where the offline dataset only contains states that are not necessarily from the same trajectories.}
    \item \textbf{Connections between reset model and trace model.} Given \hyrlo{} can be solved efficiently when a reset model is available \citep{bagnell2003policy}, we extend previous work with \emph{a reduction from the trace model to the reset model setting} via an admissibility condition where the offline distribution is realizable by the policy class. Further, we demonstrate evidence that suggests statistical separation between trace model and reset model if the admissibility condition is violated.
    \item \textbf{Efficient algorithm.} We provide the first provably efficient algorithm for \hyrlo{} with only trace model access, FOrward Observation-matching BAckward Reinforcement Learning (\Foobar). With the admissibility assumption, \Foobar{} requires the same order of samples as the previous algorithms \citep{kakade2002approximately,bagnell2003policy} that demand a reset model to compete with the best policy covered by the offline distribution.
    \item \textbf{General analysis.} 
     Our approach does not require the strong \emph{explicit structural assumptions} such as bilinear rank \citep{du2021bilinear} on the MDP and value function that previous hybrid RL analysis demanded~\citep{song2022hybrid,nakamoto2023cal}. Relaxing this assumption allows our algorithm and analysis to be more general and applicable to a wider range of problems.
     In addition, we identify situations where \Foobar{} succeeds under inadmissible offline data, and 
     provide algorithms and analysis under stationary settings. 
    \item \textbf{Empirical evaluation.} We perform experiments to show the effectiveness of our algorithm on two challenging benchmarks: the rich-observation combination lock \citep{misra2020kinematic} and high-dimensional robotics manipulation tasks \citep{rajeswaran2017learning}. We compare with the state-of-the-art hybrid RL algorithms and investigate the gap due to the more limited information in the offline dataset. 
\end{itemize}

\section{Related Work}
\label{sec:related}
\paragraph{Hybrid RL} Hybrid RL defines the setting where the agent has access to both offline data (usually generated by policies with a mixture of qualities) \citep{levine2020offline} and online interaction access. This learning framework has recently gained increasing interest due to its potential for efficient learning and practical values \citep{ross2012agnostic,nair2020awac,xie2021policy,song2022hybrid,lee2022offline,niu2022trust,ball2023efficient,nakamoto2023cal,wagenmaker2023leveraging,li2023reward,zhang2023policy,zhang2023policyfine,vemula2023virtues,swamy2023inverse,zhou2023offline}. Previous works follow the standard offline RL setting where the offline dataset contains the state, action, reward and next state information, and they have shown the statistical and computational benefit of the hybrid setting over pure online or offline setting. In this work, we consider a more general and challenging setting where the offline dataset only contains the state information. Many previous works have also considered this setting \citep{machado2017eigenoption,nair2018visual,schmeckpeper2020reinforcement,ma2022vip,baker2022video,seo2022reinforcement,ghosh2023reinforcement}, but the offline states are only used for representation learning in a separate pertaining stage, not for decision making via RL.  \ysedit{In addition, most of the works assume that the offline data consists of the state and next state pair \emph{collected from the same transition}.} 
Instead of a heuristic application of the offline observations, our work conducts the first theoretical study in this setting that captures the minimal properties of the offline distribution, and our proposed algorithm utilizes the offline data for decision-making directly.\loose

\paragraph{Learning from observation alone} Prior to the \hyrlo{} setting,  learning from state-only data has also been considered in other interactive decision-making problems, such as imitation learning and offline reinforcement learning. For example, Imitation from observation alone setting (\ilfo) \citep{nair2017combining,torabi2018generative,sun2019provably,smith2019avid,song2020provably,zhu2020off,radosavovic2021state} considers learning from a dataset of expert states, and with online interaction. If one does not have online access, the offline counterpart is the offline imitation learning setting, where the agent has access to two datasets: one offline state-only dataset (which has a mixture of qualities) and another expert state-only dataset \citep{kim2021demodice,ma2022versatile,yu2023offline,pirotta2023fast}. \ysedit{However, all these settings require explicitly labeled expert data, while our setting only requires unlabeled offline data that implicitly covers some good policy's trajectory.} Recently \citet{li2023mahalo} removed the expert label requirement, but the offline data is still required to contain additional action or reward information.

\paragraph{RL with reset model} In the reset model setting, one assumes the ability to reset the dynamics to any state. With a reset model, the \hyrlo{} problem can be solved using Policy Search by Dynamic Programming (\Psdp) algorithm and others that share a similar core idea \citep{bagnell2003policy,salimans2018learning,uchendu2023jump}.  The reset model has been shown to have other favorable properties that contribute to overcoming the statistical hardness of the more commonly available trace model setting \citep{amortila2022few,weisz2021query}. On the empirical side, \citet{sharma2022state} shows that if expert data is available, one can learn to reset by training a policy that brings the current policy to the expert state distribution after the rollout. In this paper, we also demonstrate the benefit of such “reset policy”. The previous work requires a non-stationary initial distribution for the reset policy due to the interleaving learning between the final policy and the rest policy. Our paper improves over the previous work by removing the non-stationarity with learning a reset policy before the “policy optimization” stage. In addition, the previous work does not apply to any non-reversible system, which restricts its application to real-world problems.

\section{Preliminaries}
\label{sec:prelim}
We consider finite horizon MDPs $\Mcal = \{\Scal, \Acal, H, R, P\}$, where $H$ is the horizon, $\Scal$ is the state space with $|\Scal| = S$, $\Acal$ is the action space with $|\Acal| = A$, $R = \{R_h: \Scal \times \Acal \to [0,1]\}_{h=1}^H$ is the reward function, $P = \{P_h: \Scal \times \Acal \to \Delta(\Scal)\}_{h=1}^H$ is the state transition distribution, and $P_0(\emptyset)$ is the initial state distribution. We denote the model $\Mcal$ as the trace model to distinguish it from the reset model that we will introduce later. Given a (potentially nonstationary) policy $\pi \in \Pi = \{\Scal \to \Delta(\Acal)\}_{h=1}^H$, define the action-value Q-function of $\pi$ at timestep $h$ as $Q_h^\pi(s_h,a_h) = \EE_{\pi,P} \sbr{\sum_{\tau=h}^H R_\tau(s_\tau,a_\tau)}$, and we define the optimal policy as $\pi^\ast$. We define the function class to estimate the Q function as $\Fcal: \{\Fcal_h: \Scal \times \Acal \to [0,H]\}_{h=1}^H$. We follow the conventional notation $d^\pi_h$ to denote either the state (or state-action) occupancy measure induced by $\pi$ at horizon $h$. 

We study the hybrid RL setting \citep{song2022hybrid}, where the agent has online interaction access to the environment, and in addition, has offline data set $\{\Dcal_h\}_{h=1}^H$.  In the canonical hybrid RL setting, each dataset $\Dcal_h$ contains tuples $\{s^n_h,a^n_h,r^n_h,s^n_{h+1}\}_{n=1}^N$, where $N$ is the size of the offline dataset. The data in $\Dcal_h$ is drawn from some distribution $\mu_h$, i.e., $s^n_h,a^n_h \sim \mu_h$: for example, $\mu_h$ can be the visitation distribution of some policy, and $r^n_h = R_h(s^n_h,a^n_h)$,  $s^n_{h+1} \sim P_h(\cdot \mid s^n_h, a^n_h)$. Here we consider the \hyrlo{} setting, where in the offline dataset we only have the single-timestep state data. That is, the offline dataset has the form $\Dcal_h = \{s^n_h\}_{n=1}^N$, where $s_h \sim \mu_h$, and $\mu_h$ is some distribution over the states at timestep $h$. 

Following the convention in hybrid RL, the learning goal is to compete with the best policy covered by the offline distribution. For the coverage notation, in the main text, we consider the density ratio coverage for simplicity: given any policy $\pi$, we define the density ratio coverage as
$\ccov(\pi) = \min_{h \in [H]} \nbr{ \frac{d^\pi_h}{\mu_h}}_{\infty}$, where the supremum norm is over states. \footnote{We use the general density ratio coverage for simplicity of presentation for the main text, a tighter coverage similar to \citep{song2022hybrid} applies naturally but we defer it to \pref{app:structural}.}

To measure the difference between distributions, we define the Integral Probability Metric ($\IPM$) \citep{muller1997integral} distance between two distributions $\PP$ and $\QQ$:
\begin{align*}
    \IPM_{\Gcal}(\PP,\QQ) = \sup_{g \in \Gcal} \left| \int g~\mathrm{d}\PP - \int g~\mathrm{d}\QQ \right|,
\end{align*}
which is defined by the test function class $\Gcal$. For example, when $\Gcal$ is all bounded functions, the IPM recovers the Total Variation (TV) distance, and we denote the TV distance as $\|\cdot\|_{\TV}$. When $\Gcal$ is the set of all $1$-Lipschitz functions, this definition recovers the 1-Wasserstein distance.

We note the difference between the two different access models for an MDP: we will denote the canonical trace MDPs as we defined above as $\Mcal$, where one can only reset at the initial state $P_0$ and simulate traces $\tau = \{s_1,a_1,r_1,s_2,\ldots,s_H,a_H,r_H\}$, where $s_1 \sim P_0,\; a_h \sim \pi_h(s_h),\; r_h = R_h(s_h,a_h),\; s_{h+1} \sim P_h(s_h,a_h)$. We also consider the \textit{reset} access model with the ability to simulate a reward and transition from any state-action pair: at any horizon $h$, for any $s \in \Scal$, and any action $a \in \Acal$, we can query $r_h = R_h(s_h,a_h)$, and $P_h$ to get a sample $s_{h+1} \sim P_h(\cdot \mid s_h,a_h)$. We denote this reset model $\Mreset$.                                                                 

For a more streamlined presentation, we will utilize the concept of a partial policy which operates over a sequential segment of time steps, specifically $\sbr{l \dots r} \in [H]$. This is represented as $\Pi_{l:r} := \cbr{\pi : \bigcup_{h=l}^r \Scal \to \Delta(\Acal)}$. 

\section{Algorithm}
\label{sec:algorithm}
To provide an algorithm for \hyrlo{}, in this section, we first see how this problem is solved in the reset model setting. Then we will derive a reduction from the trace model setting to the reset model setting. The resulting algorithm thus will be a two-phase algorithm: in the first phase, we run a careful reduction to the reset model setting, and in the second phase, we run the reset model algorithm to find the optimal policy. 

\subsection{Backward Algorithm: PSDP}
Suppose we have a reset model $\Mreset$, then as hinted above, we can simply apply an existing algorithm: Policy Search by Dynamic Programming (\Psdp) \citep{bagnell2003policy}.
The \Psdp{} algorithm proceeds in a backward fashion: from the last timestep $H$ to the first timestep $1$, at each timestep $h$, the algorithm first samples states from the offline dataset: $s_h \sim \Dcal_h$, followed by sampling random action $a_h \sim \piunif$, resets $\Mreset$ to $s_h,a_h$, and samples $s_{h+1} \sim P_h(s_h,a_h)$. From $s_{h+1}$ the algorithm will roll out $s_{h+1}, a_{h+1}, \ldots, s_{H},a_{H} \sim \pi_{h+1:H}$ (which are already learned in the previous timesteps). Now we have samples of the return information $\sum_{\tau=h}^H R_\tau(s_\tau,a_\tau)$ for each $s_h,a_h$ we can use cost-sensitive classification to find the one-step optimal policy $\pi_h$ which maximizes the returns following $\pi_{h+1:H}$, the previous one-step optimal policies. However, directly solving the cost-sensitive classification problem is usually computationally intractable, and here we provide a value-based version of the \Psdp{} algorithm in \pref{alg:psdp}, where we use the estimated value function as a plug-in estimator to solve the cost-sensitive classification problem.

Now with \Psdp, as long as the offline distribution $\mu$ that generates the offline dataset $\Dcal$ covers some good policy's trajectory (for example, the density ratio coverage $\ccov(\pi^\ast)$ is bounded), we will show in \pref{sec:analysis} that the returned policy is close to optimal with enough online data.

\subsection{Trace to Reset}

However, in the \hyrlo{} framework, we do not have the reset model $\Mreset$ but the more realistic trace model $\Mcal$. What can we do in this case? It turns out that with the help of the offline dataset, we can learn a policy $\pif$ that induces a state distribution similar to $\mu$. Then suppose that for each timestep $h$, we have that $\|d^{\pif}_h - \mu_h\|$ is small, where $\|\cdot\|$ is some distance metric that we care about. Then to reset to $s_h \sim \mu_h$, we can instead roll in the policy $\pif$ to timestep $h$, and we will get samples $s_h \sim d^{\pif}_h$ (as if we are sampling $s_h \sim \mu_h$), and then we can proceed to run \Psdp. In other words, we can build a reset model with $\pif$. The new algorithm is summarized in \pref{alg:psdp_trace}. Note that the only change is in lines 3 and 4. We remark that a similar idea of using \Psdp{} with a roll in policy has also been explored in previous work \citep{mhammedi2023representation}, where the roll in policy is trained from reward-free exploration techniques. \ysedit{However, the goal of reward-free exploration is to ensure optimality instead of efficiency, since reward-free exploration has a similar lower bound as regular reward-based online RL \cite{jin2020reward}.} \loose

\subsection{Forward Algorithm: FAIL}
The technical problem remaining is to learn a policy $\pif$ that induces a state distribution close to $\mu$. Inspired by the idea of state-moment-matching in \ilfo{} literature, we can adapt one such algorithm, Forward Adversarial Imitation Learning (\Fail) \citep{sun2019provably}. \Fail{} learns a set of policies $\pi_{1:H}$ from $h=1$ to $H$ in sequence. At each timestep $h$, \Fail{} rolls in the previous policies $\pi_{1:h-1}$ and samples $s_{h} \sim d^\pi_{h}$. 
It then takes a random action $a_{h} \sim \piunif$ and samples $s_{h+1} \sim P_{h}(s_{h},a_{h})$. With the dataset $\{s_{h},a_{h},s_{h+1}\}$, \Fail{} solves the following minmax game by finding a one-step policy $\pi_h$ that minimizes the $\IPM$ under discriminator class $\Gcal$, between $\pi_{1:h}$ and $\mu_{h}$, which we approximate the samples $\Dcal_{h}$: \loose
\begin{align*}
    \min_{\pi_h \in \Pi_h} \max_{g \in \Gcal_h} \EE_{s_h \sim d^\pi_{h}} \sbr{ g(s_h)} - \EE_{s_h \sim \mu_h} \sbr{g(s_{h})},
\end{align*}
These can be estimated by the dataset collected online at time step $h$:
\arxiv{
\begin{align*}
    \min_{\pi_h \in \Pi} \max_{g \in \Gcal_h} \left(\frac{1}{N} \sum_{n=1}^N \frac{\pi_h(a_{h}^{n} \mid s_{h}^{n})}{1/A} g(s_{h+1}^{n}) - \frac{1}{N'} \sum_{n=1}^{N'} g(s_{h+1}^{n})\right).
\end{align*}
}
\icml{
\begin{align*}
    \min_{\pi_h \in \Pi} \max_{g \in \Gcal_h} \left(\sum_{n=1}^N \frac{\pi_h(a_{h}^{n} \mid s_{h}^{n})}{N /A} g(s_{h+1}^{n}) -  \sum_{n=1}^{N'} \frac{g(s_{h+1}^{n})}{N'}\right).
\end{align*}
}
To solve the minmax game, we can use the common pattern of best response playing no-regret algorithm. We show one way to solve this in \pref{alg:minmax}. 
At the high level, the \Fail{} algorithm iteratively finds the solution of the minmax problem in a forward way, ensuring the policy induces similar state visitation distribution as the offline data on each timestep. We present the pseudocode of \Fail{} in \pref{alg:fail}.

\subsection{Forward-backward Algorithm: FOOBAR}
\label{sec:alg_foobar}
We are now ready to present the proposed algorithm, FOward Observation-matching BAckward Reinforcement learning (\Foobar). We present the pseudocode in \pref{alg:foobar}. In the forward phase, we run \Fail{} that outputs a sequence of policies $\pif_{1:H}$ whose state visitation distribution is close to the offline distribution $\mu$. Care is needed here because we have not defined the discriminator class we will use for the forward phase. It turns out that if we arbitrarily select the discriminator to be all bounded function, then a polynomial dependency on the number of states is required to imitate the offline distribution (see Theorem 3.2 of \citet{sun2019provably}), which is unfavorable given the fact that we already have an offline state dataset to imitate so such dependency should be avoidable in the case with a relatively high-quality offline dataset (e.g., not a uniform distribution over states). Indeed the dependency on the state is avoidable by a careful construction of the discriminator class based on the value function class that we use in the backward pass (\pref{eq:discriminator}). We provide the justification of such construction in \pref{app:proof_foobar}. Then in the backward phase, we run \Psdp-trace (\pref{alg:psdp_trace}) with the roll-in policy $\pif_{1:H}$. \pref{alg:psdp_trace} returns refined policies $\pib_{1:H}$, which can compete with the best policy covered by $\mu$. It's important to note that if the offline dataset consists of both sub-optimal and high-quality data, the refined policies $\pib_{1:H}$ can be \textit{dramatically} better than the initial policy $\pif_{1:H}$ learned by moment matching that is used to simulate the reset model.

\begin{algorithm}[t]
    \caption{FOward Observation-matching BAckward Reinforcement learning (\Foobar)}
    \begin{algorithmic}[1]
    \Require Offline dataset $\Dcal^{\textsf{off}}$, value function class $\Fcal$, policy class $\Pi$.
    \State Define the discriminator class $\Gcal := \{\Gcal_h\}_{h=1}^H$:  \algcommentlight{Discriminators take state as input while Q-functions take state-action as input.} 
    \begin{equation}\label{eq:discriminator}
        \Gcal_h = \cbr{\max_a f(\cdot,a) - f(\cdot,a') \mid f \in \Fcal_h, a' \in \Acal}.
    \end{equation}
    \State $\pif_{1:H} \leftarrow$ \pref{alg:fail} with input $\cbr{\Dcal^{\textsf{off}}, \Gcal, \Pi}$.
    \State $\pib_{1:H} \leftarrow$ \pref{alg:psdp_trace} with input $\cbr{\pif_{1:H}, \Fcal}$.
    \end{algorithmic}
    \label{alg:foobar}
\end{algorithm}

\section{Analysis}
\label{sec:analysis}
In this section we provide the analysis of the proposed algorithm. The overall proof strategy follows the intuition of the algorithm itself: 1) we need a certain closeness guarantee (\pref{thm:fail}) between the forward policy and the offline distribution, and the major difficulty is to ensure that such requirement is not too strong (which will result in a suboptimal sample complexity, c.r.~\pref{sec:alg_foobar}), 
but is sufficient to 2) show the guarantee of the downstream learning of the backward policy (\pref{thm:foobar}). We will start with an essential assumption on a property of the offline distribution.

\subsection{Admissibility}
We follow the definition of admissibility from \citet{chen2019information}:\loose
\begin{assumption}[Admissibility] \label{assum:admissibility}
    We assume the offline distribution $\mu$ is admissible: 
    \begin{align*}
        \exists \pi \in \Pi,\; \forall h \in [H],\;  \forall s,a \in \Scal \times \Acal,\; \mu_h(s,a) = d^\pi_h(s,a).
    \end{align*}
\end{assumption}
\pref{assum:admissibility} captures the situations where the offline data is generated by a single (possibly stochastic) policy, stitching policies (due to non-stationarity), or a mixture of such policies (Chapter 13 in \citep{sutton2018reinforcement}), which is how most offline datasets are generated in practice. 
Practically, this assumption might be violated by artificial data filtering, data augmentation or other perturbation of the offline data.

Next, we provide examples in which if \pref{assum:admissibility} fails, the problem is hard in trace model but remains easy with reset model access \citep{kakade2002approximately,bagnell2003policy}: \loose

\begin{proposition} \label{prop:hard_data}
    For any algorithm $\mathsf{Alg}$, denote the dataset collected by $\mathsf{Alg}$ as $D^{\mathsf{Alg}}$, and let $\widehat D$ denote the empirical distribution of 
    a dataset $D$. Then there exists an MDP $\Mcal$ with deterministic transition and a set of offline datasets $\{\Dcal_h\}$, with arbitrary sample size $\abr{\Dcal_h} = N \geq 2$, collected from the inadmissible offline distribution $\mu$ with constant coverage: $\max_h \nbr{\frac{d^{\pi^\ast}_h}{\widehat \Dcal_h}}_{\infty} = 2$ such that,  unless $\abr{D^{\textsf{Alg}}} = \Omega(A^H)$,
    we have 
    \begin{align*}
        \max_{D} \nbr{\widehat D_H^{\mathsf{Alg}} - \widehat \Dcal_H}_{\mathsf{TV}} \geq \frac{1}{2}.
    \end{align*}
    However, there exists an algorithm $\mathsf{Alg}^{\mathsf{reset}}$ that uses any offline dataset $D$ and reset model $\Mcal^{\mathsf{reset}}$ that returns optimal policy $\pi^\ast$ with sample complexity $O(A)$.
\end{proposition}
The above statement is about the hardness of collecting a dataset that matches the offline dataset: when the admissibility assumption is violated, in the worst case we need to collect a dataset that is exponentially large in the horizon; otherwise, our dataset does not contain at least half of the states 
in the offline dataset (at least half of which is expert states), with probability 1 (recall that the construction is within a deterministic MDP). In the next proposition, we show another hardness result that has direct implications on the performance of the learned policy.

\begin{proposition} \label{prop:hard_tv}
    For any state distribution $\mu_h$, $\forall h \in [H],$ let 
    \begin{align*}
     \pi^\mu_h := \argmin_{\pi_h \in \Pi_h} \nbr{d^\pi_h - \mu_h}_{\TV},
    \end{align*}
    i.e., the policy that induces the closest state distribution to the 
    offline distribution in TV. Then
     there exists an MDP $\Mcal$ and inadmissible offline distribution $\mu$, such that $\max_h \nbr{\frac{d^{\pi^\ast}_h}{\mu_h}}_{\infty} = 18$, i.e., the offline distribution has a constant coverage over the optimal policy but
    \begin{align*}
        \max_h \nbr{\frac{d^{\pi^\ast}_h}{d^{\tilde \pi}_h}}_{\infty} = \infty, \mathand
        \max_{h,s_h} \nbr{\frac{\pi^\ast_h(s_h)}{\tilde \pi_h(s_h)}}_{\infty} = \infty.
    \end{align*}
    I.e., the policy that minimizes the TV distance to the offline distribution does not cover some states from the optimal policy's trajectory, and the induced policy does not cover some actions that the optimal policy takes.
\end{proposition}
A direct implication of \pref{prop:hard_tv} is that, if reward 1 is assigned to the states that are not covered by the learned policy (or state only reachable from those states), and reward is 0 otherwise, then the policy that best mimics the offline distribution will have a constant gap to optimal policy. Also, similar to \pref{prop:hard_data}, the setup in \pref{prop:hard_tv} is not hard in the reset model settings. These results suggest the potential for a separation between trace and result model, but they are not equivalent to an information-theoretical lower bound. \loose
\subsection{Performance Guarantee of the Forward Algorithm}
Now we analyze \pref{alg:fail}. We start with an assumption that is a relaxation of \pref{assum:admissibility}, which is sufficient for our analysis. Note that in the construction of the previous hardness results, this relaxed assumption is still violated. 
\begin{assumption}[Admissibility in $\IPM$.]\label{assum:admissibility_ipm}
    There exists a policy $\pi$ such that, for all $h \in [H]$, $\IPM_{\Gcal_h}(d^\pi_{h}, \mu_h) = 0$, where $\Gcal_h$ is defined as in the \pref{eq:discriminator}.
\end{assumption}
Note that this assumption is weaker because $\Gcal$ is a subset of bounded functions, and \pref{assum:admissibility} implies 0 TV distance, which implies \pref{assum:admissibility_ipm}. Next, we introduce the Bellman Completeness assumption, which is also commonly made in \ilfo{} \citep{sun2019provably} and hybrid RL \citep{song2022hybrid,nakamoto2023cal}: \loose

\begin{assumption}[Completeness]\label{assum:completeness}
    For any $h \in [H]$, for any $g \in \Gcal_{h+1}$, there exists $f \in \Gcal_h$ such that $f = \Tcal_h g$, where $\Tcal_h$ is the Bellman operator with respect to the offline distribution at time $h$: $\Tcal_h g(s_h) = \EE_{a_h \sim \mu_h(s_h)}  \EE_{s_{h+1} \sim P_h(s_h,a_h)}[ g(s_{h+1})]$. That is,
    \begin{align*}
        \max_h \max_{g \in \Gcal_{h+1}} \min_{f \in \Gcal_h} \left\|f - \Tcal_h g\right\|_\infty = 0.
    \end{align*} 
\end{assumption}

Note that the previous two assumptions can both hold approximately, and here we assume that they hold exactly for simplicity. Now we can state the performance guarantee of the forward algorithm. The result is characterized in the IPM between the learned policy and the offline distribution.
\begin{theorem}[Guarantee of \pref{alg:fail}]\label{thm:fail}
    Assume \pref{assum:admissibility_ipm,assum:completeness} hold. Suppose $|\Dcal^{\mathsf{off}}| = |\Dcal^{\mathsf{on}}| = N$, then with probability $1-\delta$, the returned policy $\pif$ satisfies that, for any $h \in [H]$,
    \begin{align*}
        \IPM_{\Gcal_{h}}(d^{\pif}_{h}, \mu_{h}) \leq h\epsforward(\delta, N),
    \end{align*}
    where $\epsforward(\delta, N) :=$
    \begin{align*}
    8\sqrt{\frac{2A\log(2|\Gcal_{h}||\Pi_h|/\delta)}{N}} + \frac{16A\log(2|\Gcal_{h}||\Pi_h |/\delta)}{N} + \sqrt{\frac{A^2}{T}},
    \end{align*}
    where $T$ is the number of iterations in \pref{alg:minmax}.
\end{theorem}

This result indicates that if we have equally enough samples from both online and offline (which is one of the key features of hybrid RL), and we perform enough iterations of the minmax game, then we will have the guarantee that the learned forward policy will be close to the offline distribution under any discriminator in $\Gcal$. Note that this result does not imply that the learned policy is close to the offline distribution in a stronger sense such as TV distance, and we emphasize that such a stronger notion of closeness is not necessary for learning a policy that can compare with the best policy covered by the offline distribution.

\subsection{Performance Guarantee of FOOBAR}
With the guarantee of the forward algorithm, we can show the performance guarantee of \Foobar. Different from the analysis of the forward algorithm, whose result is to compare with the offline distribution, the final result of \Foobar{} is to compare with the performance of other policies. Therefore, following the convention common in hybrid RL literature \citep{bagnell2003policy, ross2012agnostic,xie2021policy,song2022hybrid}, we state the performance guarantee of \Foobar{} with respect to any policy that is covered by the offline distribution, i.e., we can compare with any policy $\picomp$ with $\ccov(\picomp) < \infty$.

In addition to the above offline coverage condition, since our algorithm involves function approximation (i.e., we use $\Fcal$ to estimate the value functions), we also require the following standard realizability assumption:
\begin{assumption}[Realizability]\label{assum:real}
For any deterministic policy $\pi$, $h \in [H]$, we have $Q^\pi_h \in \Fcal_h$.
\end{assumption}
\ysedit{Note that here we state the most general form of realizability assumption for simplicity. In the proof (\pref{app:proof_foobar}), we use a relaxed version where the realizability holds for a subset of policies and states.} We also assume the function class $|\Fcal|$ is finite \footnote{This is without loss of generality and we can also use $|\Fcal|$ to denote similar measures such as covering number or VC-dimension of the function class.}.
Now we are ready to present our main result: \loose
\begin{theorem}\label{thm:foobar}
    Suppose \pref{assum:admissibility_ipm,assum:completeness,assum:real} hold. Then with probability at least $1-\delta$, the returned policy $\pib_{1:H}$ from \pref{alg:foobar} with discriminator constructed from \pref{eq:discriminator}, $N^{\mathsf{for}}$ offline and forward samples, and $N^{\mathsf{back}}$ backward samples, satisfies that for any comparator policy $\picomp$ such that $\ccov(\picomp) < \infty$,
    \begin{align*}
        V^{\picomp} - V^{\pib} \leq \varepsilon,
    \end{align*}
    when 
    \begin{small}
    \begin{align*}
        &H \cdot N^{\mathsf{for}} = O\rbr{\frac{\ccov^2(\picomp)H^5A\log(|\Fcal_h||\Pi_h|/\delta)}{\varepsilon^2}}\mathand H \cdot N^{\mathsf{back}} = O\rbr{\frac{\ccov^2(\picomp)H^5A\log(|\Fcal_h|/\delta)}{\varepsilon^2}}.
    \end{align*}
    \end{small}
\end{theorem}
A few remarks are in order:
\begin{remark}[Reduction from trace to reset]
We can see that the samples required for the forward algorithm and backward algorithm are only different by a factor of $\log(|\Pi|)$. If we consider a policy class with the same expressiveness as the value function class (which is generally true in practice), i.e., $\log(|\Fcal||\Pi|) \approx 2\log(|\Fcal|)$, then our algorithm performs \emph{a reduction from the trace model setting to the reset model setting} with constant overhead.  \loose
\end{remark}

\begin{remark}[Removing explicit structural assumptions]\label{remark:structural}
    Note that our result is not specific to tabular MDPs. In fact, compared with previous hybrid RL (or online RL) analysis \citep{song2022hybrid,wagenmaker2023leveraging,nakamoto2023cal}, our analysis is agnostic to the structural complexity measure $d$ \citep{jin2020provably,du2021bilinear} of the MDPs and thus applies to any MDP with finite action space. For example, in the tabular setting where $d = SA$, our result has no 
    explicit dependency on the number of states $S$, and recall in \pref{prop:hard_data} we showed that a polynomial dependency on the state space size ($S = A^H$) is difficult to avoid without the admissibility assumption. Our result has a worse dependency on $A$ but we conjecture this is fundamental in the observation-only setting. We provide a thorough discussion on this topic in \pref{app:structural}.  \loose
\end{remark}

\begin{remark}[Significance of the discriminator class]
One might think that the positive result from \pref{thm:foobar} is a natural byproduct of the positive results from \Fail{} and \Psdp{}. However, we note that \Fail{} only guarantees to return a policy that is comparable to the behavior policy (offline distribution), but the learned policy can induce different visitation distribution from the behavior policy. Thus the guarantee to compare with any covered policy is not trivial, and this is addressed by the careful construction of the discriminator class $\Gcal$.
\end{remark}

\paragraph{Further practical considerations} Finally we state two additional results 
that will have direct implications on the practicality of the algorithm. First,
if the guarantee of \pref{thm:fail} breaks (i.e., $\max_h \IPM_{\Gcal_{h}}(d^{\pif}_{h}, \mu_{h}) = c$, 
where $c$ is not small), which may be caused by inadmissible offline data, 
violation of completeness assumption, or optimization error, we show that in \pref{app:imperfect_forward}
that \Foobar{} can still compare with \emph{the best policy covered by the forward policy}.
In \pref{sec:exp_inadm}, we verify empirically robustness of \Foobar{} against 
different levels of inadmissibility. 

Second, it might be computationally and memory expensive to perform non-stationary 
algorithms such as \Foobar{}, or limiting if the horizon of the problem is not known or fixed. 
As such, in \pref{app:stationary}, we provide algorithms and analysis in stationary 
setting but with interactive offline distribution. In the robotics simulation in \pref{sec:experiments}, we witness the practical value brought by both results: we obtain the optimal policy with a stationary backward policy, while the forward policy does not perfectly mimic the offline states. \loose

\section{Experiments}
\label{sec:experiments}

In the experiments, we analyze the following questions: (1) Does \Foobar{} still demonstrate the benefit of hybrid RL framework? For example, does it still efficiently solve exploration-heavy problems without explicit exploration? (2) How does \Foobar{} compare to the canonical hybrid RL algorithms, i.e., what is the price for the missing information in the offline dataset? (3) How does the performance compare with \Psdp{} if a reset model is available, and how robust is \Foobar{} against inadmissibility in practice?

We use the following two benchmarks: the combination lock \citep{misra2020kinematic} and the hammer task of the Adroit robotics from the D4RL benchmark \citep{fu2020d4rl}. The visualization can be found in \pref{fig:envs}. Both environments are challenging: the combination lock requires careful exploration and previous online RL algorithms require additional representation learning in addition to RL \citep{misra2020kinematic,zhang2022efficient,mhammedi2023representation} due to its high-dimensional observation space, which also poses challenges for our forward state-moment-matching algorithm. The hammer task has high-dimension state and action space and difficult success conditions. Similar to \citet{ball2023efficient}, we use the binary reward version of the environment. 

\subsection{Comparing to Hybrid RL}
\paragraph{Combination locks} In this section, we investigate the first two questions. We first provide a brief description of the combination lock environment, and more details can be found in \pref{app:lock_intro}. For our experiment, we set the horizon  $H=100$. In each timestep $h$ there are three latent states: two good states and one bad state. Taking only one correct action (out of 10 actions in total) makes the agent proceed to the good states in the next timestep, otherwise, it proceeds to the bad state, and bad states only transit to bad states. The agent receives a reward of 1 if it stays at the good states at $h=H$ so the reward signal is sparse, and random exploration requires $10^{100}$ episodes to receive a reward signal for the first time. 

\begin{figure*}
    \centering
    \includegraphics[width=0.6\textwidth]{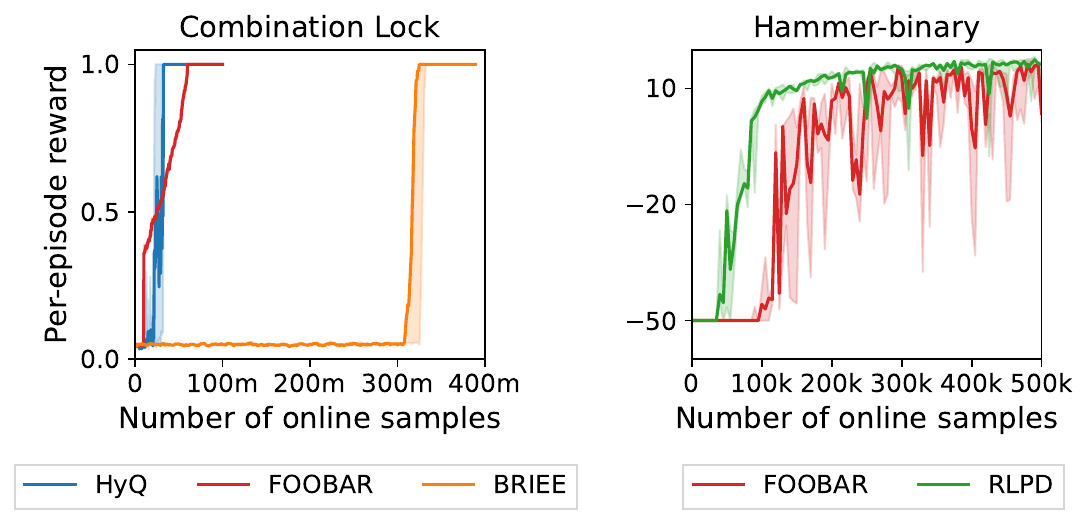}
    \caption{Comparison with hybrid RL and online RL. \textbf{Left}: evaluation curve along the training process in the combination lock task. The plot for \Foobar{} combines the forward and backward passes: during the forward pass, the evaluation result is from all the forward policies (trained and untrained). During the backward pass, after training at horizon $h$, the evaluation is from the policy $\pif \circ_h \pib$. \textbf{Right}: evaluation curve along the training process in the hammer-binary task. The plot for \Foobar{} shows the performance of the stationary backward policy in the backward phase.
    We repeat the experiment for 10 random seeds and plot the median and 25\% to 75\% percentiles.}
    \label{fig:hyrl}
\end{figure*}

We collect the offline dataset with a $\varepsilon$-greedy version of $\pi^\ast$, where $\varepsilon=\frac{1}{H}$. This guaranteed us $\ccov(\pi^\ast) \approx 2.5$. We collect 2000 samples per horizon for both \Foobar{} and Hybrid Q-Learning (\HyQ{}) \citep{song2022hybrid}, the hybrid RL algorithm that solved this task using the canonical offline dataset. We also compare with pure online RL, and we compare with the state-of-the-art algorithm in combination lock, \textsc{Briee} \citep{zhang2022efficient}. We show the result on the left of \pref{fig:hyrl}. We see that compared to the online RL method, \Foobar{} is still much more efficient, and compared with \HyQ{}, \Foobar{} indeed takes more samples but the overall sample efficiency is very comparable to the canonical hybrid RL algorithms that enjoy more information in the offline dataset. 

Our practical implementation follows the description in \pref{alg:foobar}, and we use Maximum Mean Discrepancy (MMD) \citep{gretton2012kernel} with RBF Kernel for the discriminator class, and we parameterize the policy and value functions with neural networks. We defer most implementation details to \pref{app:exp}.

\paragraph{Hammer} We also test \Foobar{} on a more popular D4RL \citep{fu2020d4rl} robotics benchmark hammer \citep{rajeswaran2017learning}. Following the evaluation protocol from \citet{ball2023efficient}, we use the binary reward version of the environment (reward of 1 if fully complete the task, -1 otherwise): thus the environment delivers sparse reward signal and the evaluation is based on how fast the agent finishes the task. To our best knowledge, we are not aware of any pure online RL method that reported solving the binary version of hammer thus we will focus our comparison to hybrid RL. We also observe that an optimal policy for this task can finish it within 50 timesteps, so we truncate the environment to $H=50$ for computational consideration.

For the implementation of \Foobar{}, we use the same implementation as the combination lock, but instead of taking random action (which is hard over the 26-dimensional continuous action space), we interleave the policy update and data collection with the latest policy. For the backward pass, we follow our stationary algorithm described in \pref{app:stationary} and use Soft Actor-Critic (SAC) \citep{haarnoja2018soft} as the policy optimization subprotocol. We present the result in \pref{fig:hyrl} (right). In the plot we only plotted the evaluation curve along the backward run for a cleaner comparison, and we use 100k samples for each horizon in the forward run (in total the forward run requires 10 times more samples than the backward run). Our backward run is comparable to \rlpd{} \citep{ball2023efficient}, but we hypothesize two reasons why our method is slightly slower: first, our forward policy does not recover the offline distribution perfectly\footnote{As we suggest in \pref{thm:foobar_for} in \pref{app:imperfect_forward}, \Foobar{} is robust to an imperfect forward run, and we provide more discussion on the empirical results in \pref{app:hammer}.}, and second, we choose to avoid some practical design choices that deviate from our theoretical algorithm but are potentially beneficial to the practical sample efficiency.
Although this is a prototypical comparison, we believe the result suggests that \hyrlo{} still demonstrates the superiority of the hybrid RL setting, but the result also suggests the gap from lacking action, reward and dynamics information in the offline dataset. Regarding the less efficient forward phase, our result in combination lock suggests that in a more controlled setting, the sample efficiency of the forward and backward run are similar (\pref{fig:zoom}), and we believe this encourages the community to design a better algorithm for state-moment-matching to close the gap further. \loose

\subsection{Inadmissible Offline Distribution}\label{sec:exp_inadm}
\begin{table}[]
    \centering
    \begin{tabular}{cccc}
    \toprule
                    & \Foobar{} (Forward) & \Foobar{} & \Psdp{} \\
    \midrule
         Benign     & $0.12~(0.1,0.135)$  & $1~(1,1)$        & $1~(1,1)$\\ 
         Adversarial &  $0~(0,0)$          & $0~(0,0)$        & 
         $1.1~(0.95,1.15)$\\
    \bottomrule
    \end{tabular}
    \caption{Comparison between \Foobar{} and \Psdp{} under inadmissible setting. We show the 
    median of the relative success rate (over optimal policy), and 25\% and 75\% percentile in the parentheses, over 10 random seeds. Note that the relative success rate in the adversarial case can exceed 1 because the environment is stochastic and the theoretical optimal policy has a success rate of 10\%. }
    \label{tab:ind}
\end{table}
To answer the last question, we construct inadmissible offline datasets in the combination lock environment with $H=10$. Specifically, we test on two inadmissible datasets: a benign dataset where the proportion of good states increases along the horizon (which is impossible for any policy to collect in a trace model setting), and an adversarial inadmissible dataset where we inject the hardness construction of \pref{prop:hard_tv} into the first horizon of the environment. We compare \Foobar{} and \Psdp{} and we present the results in \pref{tab:ind}: we can see in practice, \Foobar{} is still robust under a certain level of inadmissibility (it still solves the combination lock with the benign inadmissible dataset), but can not solve the provably hard example compared to the reset model algorithms. In this case of benign inadmissibility, the forward policy still covers the distribution of the optimal policy (the minimum coverage of good states of the offline data over the horizon is 15\%, and the forward policy has a median success rate of 12\%), leading to the final success of the whole algorithm. This again corresponds to the result of \pref{thm:foobar_for} that characterizes the success condition of \Foobar{} under inadmissibility. 

\section{Discussion}
\label{sec:discussion}
Our work initiates the theoretical study of \hyrlo,
a new theoretical paradigm with promising practical potential. 
Here we discuss some theoretical and practical open problems for future research:
\begin{itemize}
    \item Although we provide two hardness examples in the trace model setting 
    when the admissibility assumption fails, it will be interesting to understand
    if fundamental separations exist between the trace and reset model.\loose
    \item Previous hybrid RL method \citep{song2022hybrid} works under inadmissible offline distribution
    but requires structural assumption. Is there any tradeoff or connection between
    these two assumptions?
    \item Our analysis gives partial answers towards a stationary solution to 
    the \hyrlo{} problem, and it will be interesting to design a fully stationary
    algorithm for \hyrlo.
    \item Our theory suggests that a better practical implementation 
    state-moment-matching algorithm is possible and we believe this is an
    important practical problem to solve. 
\end{itemize}

\section*{Acknowledgments}
YS thanks Wen Sun for detailed feedback on the draft. The authors thank and acknowledge the support of ONR grant N000142212363 and NSF AI Institute for Societal Decision Making AI-SDM grant IIS2229881.

\icml{
\section*{Impact Statement}
This paper presents work whose goal is to advance the field of Machine Learning. There are many potential societal consequences of our work, none of which we feel must be specifically highlighted here.
}
\bibliography{refs} 
\icml{
\bibliographystyle{icml2024}
}

\clearpage

\appendix  
\onecolumn
\section{Omitted Pseudocodes}

In the following, we will utilize the concept of a partial policy (defined in the main text already but we will repeat here for completeness) which operates over a sequential segment of timesteps, specifically $\sbr{l \dots r} \in [H]$. This is represented as $\Pi_{l:r} := \cbr{\pi : \bigcup_{h=l}^r \Scal \to \Delta(\Acal)}$. Given any two intervals $1 \leq t \leq h \leq H$, we consider two partial policies: $\pi \in \Pi_{1:t-1}$ and $\pi' \in \Pi_{t:h}$. The composition $\pi \circ_t \pi'$ represents a policy that follows $\pi$ for the initial $t-1$ steps and switches to $\pi'$ for the subsequent $h-t+1$ steps. Formally, this is defined as $(\pi \circ_t \pi')(s_l) = \pi(s_l)$ when $l < t$ and $(\pi \circ_t \pi')(s_l) = \pi'(s_l)$ for $t \leq l \leq h$. The notation $s_h \sim \pi$ implies that the state $s_h$ is selected according to the distribution defined by the law of $\pi$ and $P$, and we extend this notation to include the action $a_h$ as well, denoted as $s_h, a_h \sim \pi$. 

\begin{algorithm}[h]
    \caption{Policy Search by Dynamic Programming (\Psdp)}
    \begin{algorithmic}[1]
    \Require Offline dataset $\{\Dcal^{\mathsf{off}}_h\}$, online sample size $N$.
    \For{$h=H, \ldots, 1$}
        \For{$n = 1, \ldots, N$}
            \State Sample $s^n_h \sim \Dcal^{\mathsf{off}}_h, a^n_h \sim \piunif$.
            \State Reset $\Mreset$ to $s^n_h,a^n_h$ and sample $s^n_{h+1} \sim P_h(s^n_h,a^n_h)$, $r^n_h = R_h(s^n_h, a^n_h)$.
            \State Follow $\pi_{h+1:H}$ and get sample $r^n_{h+1:H} \sim \pi_{h+1:H}.$
        \EndFor
    \State Train regressor $f_h(s_h,a_h)$ on $r_{h:H}$: \algcommentlight{Estimate Q function.}
        \begin{align*}
            f_h = \argmin_{f \in \Fcal} \sum_{n=1}^N \left(f(s_h^{n},a_h^{n}) - \sum_{\tau=h}^H r_{\tau}^{n}\right)^2.
        \end{align*}
    \State Get one-step greedy policy $\pi_h(s_h) = \argmax_{a_h} f_h(s_h,a_h)$.
    \EndFor
    \Return{Non-stationary backward policy $\pi_{1:H}$.}
    \end{algorithmic}
    \label{alg:psdp}
\end{algorithm}

\begin{algorithm}[h]
    \caption{Policy Search by Dynamic Programming (\Psdp) with trace model}
    \begin{algorithmic}[1]
        \Require Roll in policy $\pif$, online sample size $N$.
        \For{$h=H, \ldots, 1$}
            \For{n = $1, \ldots, N$}
                \State Sample $s^n_h \sim \pif_{1:h}, a^n_h \sim \piunif$.
                \State Sample $s^n_{h+1} \sim P_h(s^n_h,a^n_h)$, $r^n_h = R_h(s^n_h, a^n_h)$.
                \State Follow $\pi_{h+1:H}$ and get sample $r^n_{h+1:H} \sim \pi_{h+1:H}.$
            \EndFor
        \State Train regressor $f_h(s_h,a_h)$ on $r_{h:H}$: \algcommentlight{Estimate Q function}
            \begin{align*}
                f_h = \argmin_{f \in \Fcal} \sum_{n=1}^N \left(f(s_h^{n},a_h^{n}) - \sum_{\tau=h}^H r_{\tau}^{n}\right)^2.
            \end{align*}
        \State Get one-step greedy policy $\pi_h(s_h) = \argmax_{a_h} f_h(s_h,a_h)$.
        \EndFor
    \Return{Non-stationary backward policy $\pi_{1:H}$.}
    \end{algorithmic}
    \label{alg:psdp_trace}
\end{algorithm}

\begin{algorithm}[h]
\caption{Forward Adversarial Imitation Learning (\Fail)}
\begin{algorithmic}[1]
\Require Offline dataset $\{\Dcal^{\mathsf{off}}_h\}$, discriminator class $\Gcal=\{\Gcal_h\}_{h=1}^H$, policy class $\Pi = \{\Pi_h\}_{h=1}^H$, number of online samples N, number of iterations of minmax game $T$.
    \For{$h=1, \ldots, H$}    
    \State $\Dcal^{\mathsf{on}}_h \leftarrow \emptyset$.
    \For{$n=1, \ldots, N$}
    \State Sample $s^n_{h-1}, a^n_{h-1}, s^n_{h} \sim \pi \circ_h \piunif$.
    \State Add $(s^n_{h-1}, a^n_{h-1}, s^n_{h})$ to $\Dcal^{\mathsf{on}}_h$.
    \EndFor
    \State Get $\pi_h$ from the return of \pref{alg:minmax} with inputs $\cbr{\Pi_h, \Gcal_h, T, \Dcal^{\mathsf{on}}_h,\Dcal^{\mathsf{off}}_h}$.
    \EndFor
\Return{$\pi_{1:H}$.}
    \end{algorithmic}
    \label{alg:fail}
\end{algorithm}

\begin{algorithm}[h]
    \caption{Min-Max Game}
    \begin{algorithmic}[1]
    \Require Policy class $\Pi$, discriminator class $\Gcal$, number of iterations $T$, online dataset $\Dcal^{\mathsf{on}}$, offline dataset $\Dcal^{\mathsf{off}}$.
    \State Randomly initialize $\pi_0 \in \Pi$.
    \State Define loss function 
    \begin{align*}
        u(\pi,g) := \left(\widehat \EE_{\Dcal^{\mathsf{on}}} \sbr{\frac{\pi(a_{h-1} \mid s_{h-1})}{1/A} g(s_{h})} - \widehat \EE_{\Dcal^{\mathsf{off}}} \sbr{g(s_{h})}\right).
    \end{align*}
    \For{$t=1, \ldots, T$}
    \State $g^t = \argmax_{g \in \Gcal} u(\pi^t, g).$\algcommentlight{Linear programming oracle.}
    \State $u^t := u(\pi^t, g^t)$.
    \State $\pi^{t+1} = \argmin_{\pi \in \Pi} \sum_{\tau=1}^t u(\pi, g^t) + \phi(\pi)$. \algcommentlight{Regularized cost-sensitive oracle.}
    \EndFor
\Return $\pi^{t^\ast}$ with $t^\ast = \argmin_{t \in [T]} u^t.$
    \end{algorithmic}
    \label{alg:minmax}
\end{algorithm}

\clearpage
\section{Proof of Inadmissibility Hardness}
In this section we prove the proofs for the two hardness examples we constructed in \pref{prop:hard_data} and \pref{prop:hard_tv}.

\begin{proposition} 
    For any algorithm $\mathsf{Alg}$, denote the dataset collected by $\mathsf{Alg}$ as $D^{\mathsf{Alg}}$, and let $\widehat D$ denote the empirical distribution of 
    a dataset $D$. Then there exists an MDP $\Mcal$ with deterministic transition and a set of offline datasets $\{\Dcal_h\}$, with arbitrary sample size $\abr{\Dcal_h} = N \geq 2$, collected from the inadmissible offline distribution $\mu$ with constant coverage: $\max_h \nbr{\frac{d^{\pi^\ast}_h}{\widehat \Dcal_h}}_{\infty} = 2$ such that,  unless $\abr{D^{\mathsf{Alg}}} = \Omega(A^H)$,
    we have 
    \begin{align*}
        \max_{\Dcal} \nbr{\widehat D_H^{\mathsf{Alg}} - \widehat \Dcal_H}_{\mathsf{TV}} \geq \frac{1}{2}.
    \end{align*}
    However, there exists an algorithm $\mathsf{Alg}^{\mathsf{reset}}$ that uses any offline dataset $\Dcal$ and reset model $\Mcal^{\mathsf{reset}}$ that returns optimal policy $\pi^\ast$ with sample complexity $O(A)$.
\end{proposition}

\begin{figure}[th]
    \centering
    \includegraphics[width=0.6\textwidth]{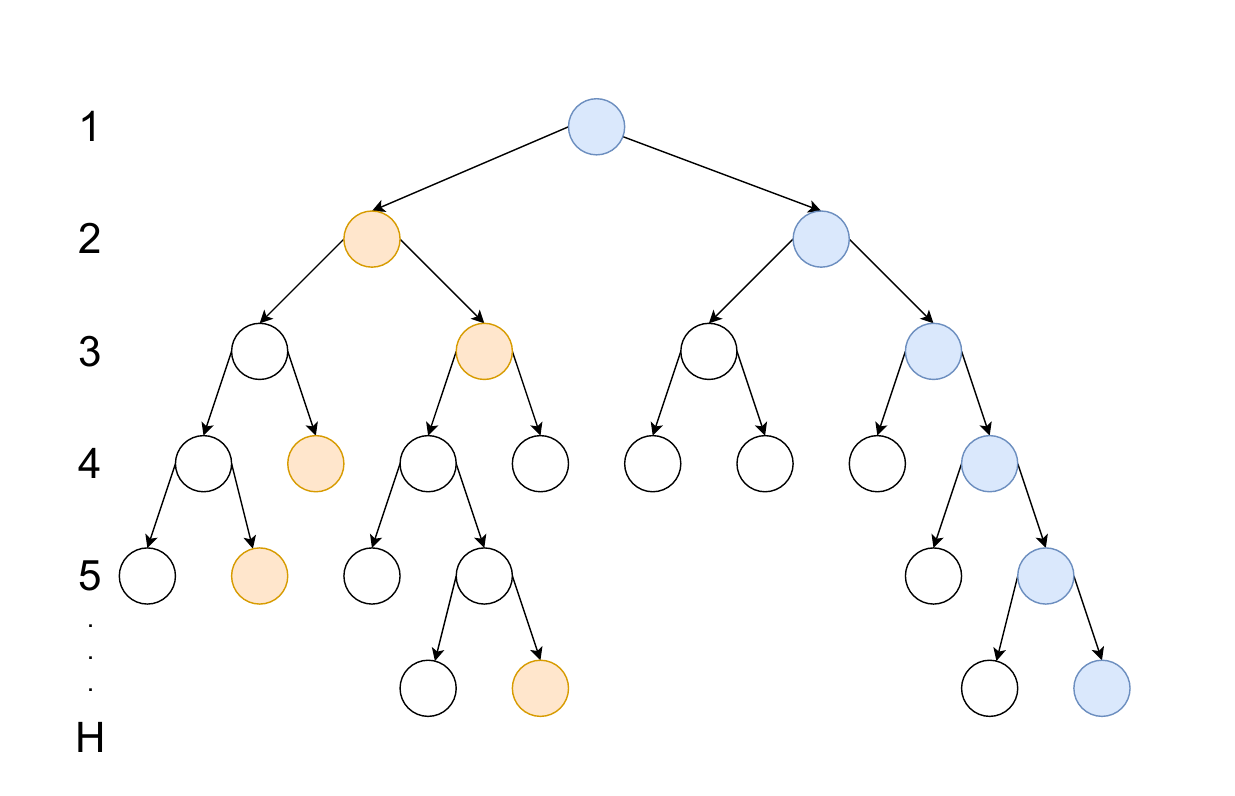}
    \caption{Construction for \pref{prop:hard_data}. The blue notes correspond to the offline data's coverage of the optimal policy. 
    The orange note corresponds to the inadmissible part of the offline data. }
    \label{fig:hard_data}
\end{figure}
\begin{proof}
    Consider a binary tree MDP $\Mcal$, with two actions and deterministic transitions. Now we construct the offline dataset $\Dcal$ as follows: for each horizon $h$, the dataset $\Dcal_h$ contains one sample from the optimal path and one sample from the other half of the tree. Thus by construction the dataset satisfies the coverage assumption. On the non-optimal data, we select them in the following way: on each level $h \geq 4$, the non-optimal data $s^{\mathsf{no}}_h$ is an arbitrary non-child node of the last horizon $s^{\mathsf{no}}_{h-1}$. Thus we see the previous non-optimal states provide no information for the current horizon, and thus the problem of finding $s^{\mathsf{no}}_{h-1}$ is equivalent to a random search over an arbitrary leaf node. However, unless $s^{\mathsf{no}}_{h}$ is added to the dataset, we will have  $\nbr{D_{h} - \Dcal_h}_{\TV} \geq \frac{1}{2}$, and thus we complete the proof. We show an example of such construction in \pref{fig:hard_data}, where the orange states denote the non-optimal states covered by the offline dataset and the blue states denote the optimal states covered by the offline dataset.
\end{proof}

\begin{proposition}
    For any state distribution $\mu_h$, let 
    \begin{align*}
     \pi^\mu_h = \argmin_{\pi_h \in \Pi} \nbr{d^\pi_h - \mu_h}_{\TV},
    \end{align*}
    i.e., the policy that induces the closest state distribution to the 
    offline distribution in TV. Then
     there exists an MDP $\Mcal$ and inadmissible offline distribution $\mu$, such that $\max_h \nbr{\frac{d^{\pi^\ast}_h}{\mu_h}}_{\infty} = 18$, i.e., the offline distribution has a constant coverage and we have
    \begin{align*}
        \max_h \nbr{\frac{d^{\pi^\ast}_h}{d^{\tilde \pi}_h}}_{\infty} = \infty, \mathand
        \max_{h,s_h} \nbr{\frac{\pi^\ast_h(s_h)}{\tilde \pi_h(s_h)}}_{\infty} = \infty.
    \end{align*}
    i.e., the policy that minimizes the TV distance to the offline distribution does not cover some states from the optimal policy's trajectory, and the induced policy does not cover some actions that the optimal policy takes.
\end{proposition}

\begin{figure}[th]
    \centering
    \includegraphics[width=0.5\textwidth]{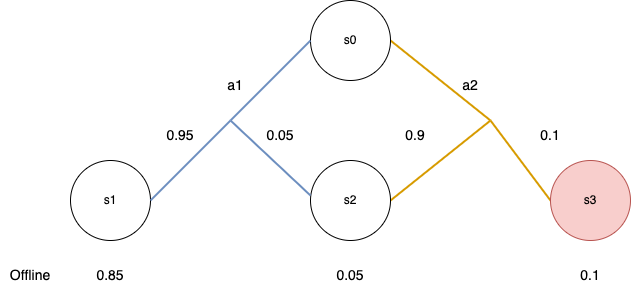}
    \caption{Construction for \pref{prop:hard_tv}. The blue transition corresponds to the dynamics after taking the action $a_1$, and the orange transition corresponds to the dynamics after taking the action $a_2$. The red node denotes the node with rewards.}
    \label{fig:hard_tv}
\end{figure}
\begin{proof}
    Consider the following one-step transition: where we start from $s_0$, and action $a_1,a_2$, and next states $s_1,s_2, s_3$, We have the following transition: $P(\cdot \mid s_0, a_1) = [0.95, 0.05, 0]$ and $P(\cdot \mid s_0, a_2) = [0, 0.9, 0.1]$. Consider the inadmissible offline state distribution over $s_1,s_2,s_3: \mu = [0.85,0.05,0.1]$. We can verify that the coverage assumption holds. Also suppose $\pi^\ast(s_0) = a_2$. Then by some calculation we have $\pi^{\mu}(s_0) = a_1$, and thus we have $\nbr{\frac{d^{\pi^\ast}}{d^{\tilde \pi}}}_{\infty} = \infty$, i.e., $\tilde \pi$ does not visit $s_3$ but $\pi^\ast$ does. An illustration of the construction can be found in \pref{fig:hard_tv}.
\end{proof}

\clearpage
\section{Proof of FOOBAR}

\subsection{Proof of \pref{thm:fail}}
For the proof, we define a shorthand notation for the IPM distance as follows:
\begin{align*}
    d_{\Gcal_{h}}(\pi \mid \rho_{h-1}, \mu_{h}) := \IPM_{\Gcal_{h}}(\rho \circ_h \pi, \mu_{h}).
\end{align*}
Then we have the following guarantee of the \pref{alg:minmax}:
\begin{lemma}[Guarantee of \pref{alg:minmax} (Theorem 3.1 of \citet{sun2019provably})] \label{lem:minmax}
Assume \pref{assum:admissibility_ipm} holds. Suppose that $\Gcal_{h}$ is the discriminator class, $\rho$ is the roll in distribution, and $\mu$ is the offline distribution. Let $|\Dcal^{\mathsf{on}}| = |\Dcal^{\mathsf{off}}| = N$, then with probability $1-\delta$, the returned policy $\pi$ satisfies:
\begin{align*}
    d_{\Gcal_{h}}(\pi \mid \rho_{h-1}, \mu_{h}) \leq \min_{\pi' \in \Pi}d_{\Gcal_{h}}(\pi' \mid \rho_{h-1}, \mu_{h}) + \epsforward(\delta, N),
\end{align*} 
where 
\begin{align*}
    \epsforward(\delta, N) = 8\sqrt{\frac{2A\log(2|\Gcal_{h}||\Pi|/\delta)}{N}} + \frac{16A\log(2|\Gcal_{h}||\Pi|/\delta)}{N} + \sqrt{\frac{A^2}{T}}.
\end{align*}
\end{lemma}
The proof of \pref{lem:minmax} can be found in \citet{sun2019provably}. The result is the standard concentration argument and taking union bound over the discriminator class and policy class, in addition to 
the no-regret guarantee of \pref{alg:minmax}.

Now we can prove the guarantee of \pref{alg:fail} in IPM, which is the property required in proving the final result.

\begin{theorem}[Restatement of \pref{thm:fail}]
Assume \pref{assum:completeness} and \pref{lem:minmax} hold. Suppose $|\Dcal^{\mathsf{on}}| = |\Dcal^{\mathsf{off}}| = N$, then with probability $1-\delta$, the returned policy $\pif$ satisfies that, for any $h \in [H]$,
\begin{align*}
    \IPM_{\Gcal_{h}}(\pif_{1:h}, \mu_{h}) \leq h\epsforward(\delta, N),
\end{align*}
where $\epsforward(\delta, N) :=$
    \begin{align*}
    8\sqrt{\frac{2A\log(2|\Gcal_{h}||\Pi_h|/\delta)}{N}} + \frac{16A\log(2|\Gcal_{h}||\Pi_h |/\delta)}{N} + \sqrt{\frac{A^2}{T}},
    \end{align*}
    where $T$ is the number of iterations in \pref{alg:minmax}.
\end{theorem}
\begin{proof}
    We can prove by induction. Consider the $h$ timesteps, where we have 
    \begin{align*}
        \IPM_{\Gcal_{h-1}}(\pif_{1:h-1}, \mu_{h-1}) \leq (h-1)\epsforward(\delta, N).
    \end{align*}
    Let $\pi^\ast := \argmin_{\pi \in \Pi} d_{\Gcal_h}(\pi \mid \pif_{1:h-1}, \mu_h)$, we have:
    \begin{align*}
        \IPM_{\Gcal_{h}}(\pif_{1:h}, \mu_{h}) &\leq d_{\Gcal_h}(\pi^\ast \mid \pif_{1:h-1}, \mu_h) + \epsforward(\delta, N) \tag{\pref{lem:minmax}} 
        \\&= \max_{g \in \Gcal_h}\abr{\EE_{s_{h} \sim \pi_{h-1}, a_{h} \sim \pi^\ast(s_{h}), s_{h+1} \sim P_h(s_h,a_h)}[g(s_{h+1})] - \EE_{s_{h} \sim \mu_{h}, a_{h} \sim \mu(s_{h}), s_{h+1} \sim P_h(s_h,a_h)}[g(s_{h+1})}]
        \\&\hspace{0.75\textwidth}+ \epsforward(\delta, N) 
        \\&\leq \max_{g \in \Gcal_h}\abr{\EE_{s_{h} \sim \pi_{h-1}, a_{h} \sim \mu(s_{h}), s_{h+1} \sim P_h(s_h,a_h)}[g(s_{h+1})] - \EE_{s_{h} \sim \mu_{h}, a_{h} \sim \mu(s_{h}), s_{h+1} \sim P_h(s_h,a_h)}[g(s_{h+1})]} \\
        &\hspace{0.75\textwidth} + \epsforward(\delta, N).
    \end{align*}

    Now denote $g^\ast_{h} = \argmax_{g \in \Gcal_h}\abr{\EE_{s_{h} \sim \pi_{h-1}, a_{h} \sim \mu(s_{h}), s_{h+1} \sim P_h(s_h,a_h)}[g(s_{h+1})] - \EE_{s_{h} \sim \mu_{h}, a_{h} \sim \mu(s_{h}), s_{h+1} \sim P_h(s_h,a_h)}[g(s_{h+1})]}$, we denote 
    \begin{align*}
        g^\ast_{h-1} = \argmin_{g \in \Gcal_{h-1}} \nbr{g - \Tcal_{h-1} g^\ast_{h}}_\infty,
    \end{align*}
    then we have 
    \begin{align*}
        &\abr{\EE_{s_{h} \sim \pi_{h-1}, a_{h} \sim \mu(s_{h}), s_{h+1} \sim P_h(s_h,a_h)}[g^\ast_h(s_{h+1})] - \EE_{s_{h} \sim \mu_{h}, a_{h} \sim \mu(s_{h}), s_{h+1} \sim P_h(s_h,a_h)}[g^\ast_h(s_{h+1})]} \\\leq& \abr{\EE_{s_{h} \sim \pi_{h-1}}[g^\ast_{h-1}(s_{h})] - \EE_{s_{h} \sim \mu_{h}}[g^\ast_{h-1}(s_{h})]} + \varepsilon^{\mathsf{be}} \tag{\pref{assum:completeness}} \\
        \leq& \IPM_{\Gcal_{h-1}}(\pif_{1:h-1}, \mu_{h-1}) + \varepsilon^{\mathsf{be}}  \\
        \leq& (h-1)\epsforward(\delta, N) + \varepsilon^{\mathsf{be}}, \tag{Inductive hypothesis}
    \end{align*}
    and thus we complete the proof.
\end{proof}

\subsection{Proof of \pref{thm:foobar}.}
\label{app:proof_foobar}

Now we prove the guarantee of \pref{alg:foobar}. We start with the guarantee of the value function estimation in \pref{alg:psdp_trace}.
\begin{lemma}\label{lem:concentration}
    Suppose \pref{assum:real} holds. For any $h \in [H]$, let $f_h$ be the returned value function from running \pref{alg:psdp_trace} With $|\Don_h| = N$, then with probability at least $1-\delta$, we have:
    \begin{align*}
        \EE_{s_h \sim \pif_{1:h-1}} \max_{a_h} \left[ \left(f_h(s_h,a_h) - Q^{\pib}_h(s_h,a_h)\right)^2 \right] \lesssim 
        \frac{H^2 A \log(|\Fcal|/\delta)}{n} := \rbr{\epsback(\delta, N)}^2.
    \end{align*}
\end{lemma}

\begin{proof}
First we have 
\begin{align*}
    \EE_{s_h \sim \pif_{1:h-1}} \max_{a_h} \left[ \left(f_h(s_h,a_h) - Q^{\pib}_h(s_h,a_h)\right)^2 \right] &\leq
     \EE_{s_h \sim \pif_{1:h-1}} \left[\sum_{a \in \Acal}  \left(f_h(s_h,a) - Q^{\pib}_h(s_h,a)\right)^2 \right] \\
     &= A\EE_{s_h, a_h \sim  \pif \circ_h \piunif}  \left[ \left(f_h(s_h,a_h) - Q^{\pib}_h(s_h,a_h)\right)^2 \right].
\end{align*}
Then follows standard least-square analysis (Lemma A.11, \citet{agarwal2019reinforcement}), since $\pib \circ_h \piunif$ is our roll-in distribution, we have 
\begin{align*}
    \EE_{s_h, a_h \sim  \pif \circ_h \piunif}  \left[ \left(f_h(s_h,a_h) - Q^{\pib}_h(s_h,a_h)\right)^2 \right] \lesssim \frac{H^2 \log(|\Fcal|/\delta)}{N},
\end{align*}
where $H$ is the range of the regression target. 
\end{proof}

Recall our construction of the discriminator class $\Gcal_h$:
\begin{align*}
    \Gcal_h = \cbr{\max_a f(\cdot,a) - f(\cdot,a') \mid f \in \Fcal_h, a' \in \Acal}.
\end{align*}
The reason for such construction will be clear in the proof of \pref{thm:foobar}. But we first show that the size of the discriminator class is bounded by the size of the value function class so it is not big. We assume that the value function class is finite for simplicity, but the results can be easily extended to the infinite case.
\begin{lemma}\label{lem:discriminator_size}
    $|\Gcal_h| \leq |\Fcal_h||\Acal|$.
\end{lemma}
The proof follows immediately from the construction of the discriminator class. 

Then we can show the performance guarantee of \pref{alg:foobar}:
\begin{theorem}[Restatement of \pref{{thm:foobar}}]
    Suppose \pref{assum:admissibility_ipm,assum:completeness,assum:real} hold. Then with probability at least $1-\delta$, the returned policy $\pib_{1:H}$ from \pref{alg:foobar} with discriminator constructed from \pref{eq:discriminator}, $N^{\mathsf{for}}$ offline and forward samples, and $N^{\mathsf{back}}$ backward samples, satisfies that for any comparator policy $\picomp$ such that $\ccov(\picomp) < \infty$,
    \begin{align*}
        V^{\picomp} - V^{\pib} \leq \varepsilon,
    \end{align*}
    when 
    \begin{align*}
        N^{\mathsf{for}} = O\rbr{\frac{\ccov^2(\picomp)H^4A\log(|\Fcal||\Pi|/\delta)}{\varepsilon^2}},~~ N^{\mathsf{back}} = O\rbr{\frac{\ccov^2(\picomp)H^4A\log(|\Fcal|/\delta)}{\varepsilon^2}}.
    \end{align*}
\end{theorem}

\begin{proof}
    By performance difference lemma \citep{kakade2002approximately}, we have that 
    \begin{align*}
        &~~~~V^{\picomp} - V^{\pib_{1:H}} \\&= \sum_{h=1}^H \EE_{s_h,a_h \sim d^{\picomp}_h} \left[  Q^{\pib}_h(s_{h},a_h) - Q^{\pib}_h(s_{h},\pib_h(s_{h}))  \right] \\
        &\leq  \sum_{h=1}^H \EE_{s_h \sim d^{\picomp}_h} \left[  \max_a Q^{\pib}_h(s_{h},a) - Q^{\pib}_h(s_{h},\pib_h(s_{h}))  \right] \\
        &\leq \ccov(\picomp)\sum_{h=1}^H \EE_{s_h \sim \mu_h} \left[  \max_a Q^{\pib}_h(s_{h},a) - Q^{\pib}_h(s_{h},\pib_h(s_{h}))  \right] \tag{Non-negativity}\\
        &\leq \ccov(\picomp) \rbr{\sum_{h=1}^H \EE_{s_h \sim d^{\pif}_h} \left[  \max_a Q^{\pib}_h(s_{h},a) - Q^{\pib}_h(s_{h},\pib_h(s_{h}))  \right] 
        + \IPM_{\Gcal_h}\rbr{d^{\pif}_h || \mu_h} \tag{Construction of $\Gcal$}} \\
        &\leq  \ccov(\picomp) \cdot \\&~~ \rbr{\sum_{h=1}^H \EE_{s_h \sim d^{\pif}_h} \left[  \left|\max_a Q^{\pib}_h(s_{h},a) -f_h(s_h,\pib_h(s_{h})) \right|+ \left|f_h(s_h,\pib_h(s_{h})) - Q^{\pib}_h(s_{h},\pib_h(s_{h}))\right|  \right] + \IPM_{\Gcal_h}\rbr{d^{\pif}_h || \mu_h}}\\
        &\leq  \ccov(\picomp) \cdot \\ &~~ \rbr{\sum_{h=1}^H \EE_{s_h \sim d^{\pif}_h} \left[  \max_a \left|Q^{\pib}_h(s_{h},a) - f_h(s_h,a) \right|+ \left|f_h(s_h,\pib_h(s_{h})) - Q^{\pib}_h(s_{h},\pib_h(s_{h}))\right|  \right] + \IPM_{\Gcal_h}\rbr{d^{\pif}_h || \mu_h}} \tag{$\max_x|f(x) - g(x)| \geq |\max_x f(x)-\max_x g(x)|$}\\
        &\lesssim  \ccov(\picomp)H \epsback(\delta, N^{\mathsf{back}}) + \ccov(\picomp)H^2 \epsforward(\delta,  N^{\mathsf{for}}). \\
    \end{align*}
    The last step is by Jensen's inequality, \pref{lem:concentration} and \pref{thm:fail}. By plugging in the definition of $\epsback$ and $\epsforward$, we have 
    \begin{align*}
        V^{\picomp} - V^{\pib_{1:H}} \leq O\rbr{\ccov(\picomp) H \sqrt{\frac{H^2A\log(|\Fcal|/\delta)}{N^{\mathsf{back}}}} + \ccov(\picomp) H \sqrt{\frac{A\log(|\Fcal||\Pi|/\delta)}{N^{\mathsf{for}}}} },
    \end{align*}
    by setting $T = AN^{\mathsf{for}}$ and by \pref{lem:discriminator_size}, finally by setting $N$ properly, we have the desired result.
\end{proof}

Here we remark that in order for our proof to hold, we only require the following weaker notion of realizability of value function class $\Fcal$: we only require that $Q^{\pib} \in \Fcal$, and in fact we only require it to hold under state visited by $\pif$ and $\mu$.

\clearpage
\section{Practical Considerations}
\subsection{Imperfect Forward Run}\label{app:imperfect_forward}
One advantage of this forward-backward algorithm is that, either due to optimization error or insufficient data size, if $\epsforward$ is not small, we can still guarantee the final performance as long as the following holds: define the coverage with respect to the forward policy as 
\begin{align*}
    \ccov^{\mathsf{for}}(\pi) := \max_h \nbr{\frac{d^\pi_h}{d^{\pif}_h}}_{\infty}.
\end{align*}
and we can have the following guarantee:
\begin{theorem}[\Foobar{} guarantee for imperfect forward run]\label{thm:foobar_for} 
    With probability at least $1-\delta$, 
    for any comparator policy $\picomp$, we have
    \begin{align*}
        V^{\picomp} - V^{\pib_{1:H}} \leq O\rbr{\ccov^{\mathsf{for}}(\picomp) H 
        \sqrt{\frac{H^2A\log(|\Fcal|/\delta)}{N^{\mathsf{back}}}}}.
    \end{align*}
\end{theorem}
Note that the number of offline and online forward samples will contribute to the 
term $\ccov^{\mathsf{for}}(\picomp)$, but here we make their relationship implicit.
The theorem states that by paying $\ccov^{\mathsf{for}}(\picomp)$ (but potentially $\ccov^{\mathsf{for}}(\picomp)
> \ccov(\picomp)$),
we can avoid paying the additive term 
$\ccov(\picomp) H \sqrt{\frac{A\log(|\Fcal||\Pi|/\delta)}{N^{\mathsf{for}}}}$. The practical 
application of this theorem can be demonstrated by our result for the robotics 
task in \pref{sec:experiments}, where perfectly mimicking the offline distribution 
is hard due to the high-dimensional continuous action space. Nevertheless, the forward
policy still covers the optimal policy, and thus \Foobar{} returns the optimal policy 
after the backward phase. We show the proof below:
\begin{proof}
    We have again by performance difference lemma,
    \begin{align*}
        &~~~~V^{\picomp} - V^{\pib_{1:H}}  
        \\&= \sum_{h=1}^H \EE_{s_h,a_h \sim d^{\picomp}_h} \left[  Q^{\pib}_h(s_{h},a_h) - Q^{\pib}_h(s_{h},\pib_h(s_{h}))  \right] \\
        &\leq  \sum_{h=1}^H \EE_{s_h \sim d^{\picomp}_h} \left[  \max_a Q^{\pib}_h(s_{h},a) - Q^{\pib}_h(s_{h},\pib_h(s_{h}))  \right] \\
        &\leq  \sum_{h=1}^H \EE_{s_h \sim d^{\picomp}_h} \left[  \max_a \left|Q^{\pib}_h(s_{h},a) - f_h(s_h,a) \right|+ \left|f_h(s_h,\pib_h(s_{h})) - Q^{\pib}_h(s_{h},\pib_h(s_{h}))\right|  \right] \\
        &\leq  \ccov^{\mathsf{for}}(\picomp) \rbr{\sum_{h=1}^H \EE_{s_h \sim d^{\pif}_h} \left[  \max_a \left|Q^{\pib}_h(s_{h},a) - f_h(s_h,a) \right|+ \left|f_h(s_h,\pib_h(s_{h})) - Q^{\pib}_h(s_{h},\pib_h(s_{h}))\right|  \right] } \\
        &\leq  \ccov^{\mathsf{for}}(\picomp)H \epsback(\delta, N^{\mathsf{back}}),
    \end{align*}
    note that in this case, we can directly shift the distribution from $\picomp$ to 
    $\pif$ in line 3. The rest of the proof is the same as the proof of \pref{thm:foobar}.
\end{proof}

\subsection{Stationary Results}\label{app:stationary}
In this section we introduce a variant of our algorithm and analysis in the stationary setting. We start with 
introducing new notations for the stationary setting, and like the procedure in our main text, we will first introduce the backward phase of the algorithm, which requires no additional assumptions and directly extends to the stationary setting. We end up with our forward phase algorithm, which requires an additional assumption that the offline dataset is interactive.

\paragraph{Notations} We start with introducing the notations in the stationary setting. In the stationary setting, 
we are interested in the finite horizon discounted MDP $\Mcal = (\Scal, \Acal, P, R, \gamma)$, 
where the transition kernel $P: \Scal \times \Acal \to \Delta(\Scal)$ and reward function $R:
\Scal \times \Acal \to [0,1]$ are stationary. We denote $\gamma$ as the discounted factor. 
For any policy $\pi$, we denote the value function as $V^\pi(s) = \EE_{\pi,P} \sbr{\sum_{h=1}^\infty \gamma^h R(s_h,a_h)\mid s_1 = s}$ and $Q^\pi(s) = \EE_{\pi,P} \sbr{\sum_{h=1}^\infty \gamma^h R(s_h,a_h)\mid s_1 = s,a_1 = a}$. We still denote $d^\pi_h$ as the visitation distribution of policy 
$\pi$ at horizon $h$, and we will often use $d^\pi = (1-\gamma)\sum_{h=1}^\infty \gamma^h d^\pi_h$ 
as the stationary visitation distribution (or occupancy measure) of policy $\pi$. We denote $\mu$ as the offline distribution that is 
constructed in a similar manner, i.e., $\mu = (1-\gamma)\sum_{h=1}^\infty \gamma^h \mu_h$, and 
we denote the coverage of $\mu$ as $\ccov(\pi) = \nbr{\frac{d^{\pi}}{\mu}}_{\infty}$. Finally in 
this section, to simply the notation, we will make extensive use of the notion of advantage, which is defined as 
$A^{\pi}(s,\pi'(s)) = \EE_{a \sim \pi'(s)} [Q^{\pi}(s,a) - Q^{\pi}(s,\pi(s))].$

\begin{algorithm}[t]
    \caption{Conservative Policy Iteration (\Cpi) with trace model}
    \begin{algorithmic}[1]
        \Require Roll in policy $\pif$, accuracy parameter $\varepsilon$, step size $\alpha$.
        \State Initialize $\pi^1$ randomly.
        \For{$t = 1, 2, \ldots$}
        \State $\pi' \gets \mathsf{Greedy}_{\varepsilon}(\pi^{t}, \Pi,d^{\pif})$.
        \If{$\EE_{s \sim \pi^t}A^{\pi^t}(s,\pi'(s)) \leq \varepsilon$}
        \State \Return $\pi^t$.
        \EndIf
        \State Update policy conservatively:
        \begin{equation}\label{eq:conservative_update}
            \pi^{t+1} \gets (1-\alpha) \pi^t + \alpha \pi'.
        \end{equation}
        \EndFor
    \end{algorithmic}
    \label{alg:cpi_trace}
\end{algorithm}

\paragraph{Backward phase} For the backward phase, we will use the classic Conservative 
Policy Iteration (\Cpi{}) \citep{kakade2002approximately} algorithm, which is a stationary algorithm that guarantees the 
optimality of the returned policy under an exploratory reset distribution, which in 
our case will be our forward policy's state visitation distribution. The intuition of 
\Cpi{} is similar to the backward pass of \Foobar{}, where we first roll in the 
forward policy $\pif$, and then we will update our backward policy by rolling out and 
perform policy optimization. Specifically, given a policy $\pi$, a policy class
$\Pi$ and an initial distribution $\mu$, the output of the greedy policy selector 
$\pi' \gets \mathsf{Greedy}_{\varepsilon}(\pi,\Pi,\mu)$ has the following guarantee:
\begin{align*}
    \EE_{s \sim d^{\pi}_\mu} [A^{\pi}(s,\pi'(s))] \geq \max_{\widetilde \pi \in \Pi}
    \EE_{s \sim d^{\pi}_\mu} [A^{\pi}(s,\widetilde \pi(s))] - \varepsilon,
\end{align*}
where $d^\pi_\mu$ is the state visitation distribution of policy $\pi$ under the
initial distribution $\mu$. In practice, to ensure that the initial distribution 
of the policy optimization problem is $d^{\pif}$, we can start to roll in $\pif$,
and at each timestep, we will start to switch to roll out policy $\pi$ with probability
$1-\gamma$ \citep{agarwal2020optimality}.
In the stationary setting, however, we do not use the greedy policy 
as the next policy, because  
we can not guarantee the optimality in an inductive way, but we can still ensure
a local improvement by performing a conservative policy update (\pref{eq:conservative_update}). 
We provide the pseudocode of \Cpi{} in \pref{alg:cpi_trace}. 

For simplicity, we will not perform the finite sample analysis on the $\mathsf{Greedy}_{\varepsilon}$
subprocedure, but we will assume that the greedy policy selector guarantee always holds, 
and we will prove the final optimality result based on it. We first state a critical lemma that 
is useful for the analysis of \Cpi{}:
\begin{lemma}[Local optimality of \Cpi{}, Theorem 14.3 of \citet{agarwal2019reinforcement}]\label{lem:cpi_local}
\pref{alg:cpi_trace} terminates in at most $8\gamma/\varepsilon^2$ steps and the output 
policy $\pi$ satisfies that
\begin{align*}
    \max_{\pi' \in \Pi} \EE_{s \sim d^{\pi}_{\mu}} [A^{\pi}(s,\pi'(s))] \leq 2\varepsilon,
\end{align*}
where $\mu = d^{\pif}$.
\end{lemma}

With the local optimality guarantee, we can show that the result for \pref{alg:cpi_trace}:
\begin{theorem}[Guarantee of \Cpi-trace]\label{thm:cpi_trace}
    Let the returned policy of \pref{alg:cpi_trace} be $\pib$, suppose policy class $\Pi$
    is realizable in the sense that 
    \begin{align*}
        \EE_{s \sim d^{\pib}} \sbr{\max_{a \in \Acal} A^{\pib}(s,a)} - \EE_{s \sim d^{\pib}} \sbr{A^{\pib}(s,\pi(s))} = 0.
    \end{align*}
    Then we have that
    \begin{align*}
        V^{\picomp} - V^{\pib} \leq \frac{\ccov(\picomp)}{(1-\gamma)^2}(2\varepsilon) + \frac{\ccov(\picomp)}{(1-\gamma)}\IPM_{\Gcal}\rbr{d^{\pif}_h || \mu_h}.
    \end{align*}
\end{theorem}
\begin{proof}
    By performance difference lemma \citep{kakade2002approximately}, we have that
    \begin{align*}
        V^{\picomp} - V^{\pib} &= \frac{1}{1-\gamma} \EE_{s \sim d^{\picomp}} \sbr{A^{\pib}(s,\picomp(s))} \\ &\leq \frac{1}{1-\gamma} \EE_{s \sim d^{\picomp}} \sbr{\max_{a \in \Acal} A^{\pib}(s,a)} \\
        &\leq \frac{1}{1-\gamma} \ccov(\picomp) \EE_{s \sim \mu} \sbr{\max_{a \in \Acal} A^{\pib}(s,a)} \\
        &\leq \frac{1}{1-\gamma} \ccov(\picomp) \rbr{\EE_{s \sim \pif} \sbr{\max_{a \in \Acal} A^{\pib}(s,a)}
        + \IPM_{\Gcal}\rbr{d^{\pif} || \mu}} \\
        &\leq \frac{\ccov(\picomp)}{(1-\gamma)^2} \EE_{s \sim d^{\pib}} \sbr{\max_{a \in \Acal} A^{\pib}(s,a)}
        + \frac{\ccov(\pib)}{(1-\gamma)}\IPM_{\Gcal}\rbr{d^{\pif} || \mu} \tag{$d^{\pib}(s) \geq (1-\gamma)\pif(s)$}\\
        &\leq \frac{\ccov(\pib)}{(1-\gamma)^2} \EE_{s \sim d^{\pib}} \sbr{
            \max_{pi' \in \Pi} \EE_{s \sim d^{\pib}}\sbr{A^{\pib}(s,\pi'(s))} - 
            \max_{pi' \in \Pi} \EE_{s \sim d^{\pib}}\sbr{A^{\pib}(s,\pi'(s))} +
            \max_{a \in \Acal} A^{\pib}(s,a)
        }  \\
        &~~~+ \frac{\ccov(\pib)}{(1-\gamma)}\IPM_{\Gcal}\rbr{d^{\pif}_h || \mu_h} \\
        & \leq \frac{\ccov(\pib)}{(1-\gamma)^2}(2\varepsilon) + \frac{\ccov(\pib)}{(1-\gamma)}\IPM_{\Gcal}\rbr{d^{\pif} || \mu} \tag{\pref{lem:cpi_local} and realizability}.
    \end{align*}
    Note that for the simplicity of the presentation, we denote $d^{\pib} := d^{\pib}_\mu$ where 
    $\mu := d^{\pif}$. 
\end{proof}
\pref{thm:cpi_trace} states that, as long as the forward policy $\pif$ covers the comparator policy, 
we can guarantee the performance of the returned policy $\pib$ is close to the best comparator policy.
Next we see how we can achieve the guarantee of the forward policy in the stationary setting.

\paragraph{Forward phase} 
In the forward phase, we assume that we have an interactive offline distribution $\muit$, which for any state $s$, if we 
query the offline distribution $\muit$ with $s$, we will return a sample $s'$ by $a \sim \muit(s), s' \sim P(s,a)$. However, since we are in the observation-only setting, we only observe $s'$ but not $a$, and thus this is a relaxation from the previous works that assume interactive experts which also provide the action information \citep{ross2011reduction,ross2012agnostic}.  

Our roll-in procedure is similar to the backward phase, where for each horizon, we will have probability $(1-\gamma)$ to terminate the roll in on that horizon. Denote the state at termination as $s$, we will take a random action $a \sim \piunif$ and observe $s'$, and we add the tuple $(s,a,s')$ to the online dataset $\Don$, and similarly, we query $\muit$ with $s$, and get $s' \sim \muit(s)$, and add $(s')$ to offline dataset $\Doff$.
Then we perform the best-response playing no-regret algorithm to iteratively update our policy, similar to \pref{alg:minmax}. The full pseudocode is in \pref{alg:fail_inter}.

\begin{algorithm}[htp]
\caption{Interactive Forward Adversarial Imitation Learning (Inter-\Fail)}
\begin{algorithmic}[1]
\Require Discriminator class $\Gcal$, policy class $\Pi$, number of iterations $T$.
    \State $\Don \gets \emptyset, \Doff \gets \emptyset$.
    \State Randomly initialize $\pi^1$.
    \For{$t=1$ to $T$}    
    \State Sampling stopping time $h \sim \mathsf{Geom}(1-\gamma)$.
    \State Sample $s,a,s' \sim \pi^{t} \circ_h \piunif$, and add $(s,a,s')$ to $\Don$.
    \State Sample $s' \sim \mu(s)$, and add $s'$ to $\Doff$.
    \begin{align*}
        u(\pi,g) := \left(\widehat \EE_{\Dcal^{\mathsf{on}}} \sbr{\frac{\pi(a \mid s)}{1/A} g(s')} - \widehat \EE_{\Dcal^{\mathsf{off}}} \sbr{g(s')}\right).
    \end{align*}
    \State $g^t = \argmax_{g \in \Gcal} u(\pi^t, g).$\algcommentlight{Linear programming oracle.}
    \State $u^t := u(\pi^t, g^t)$.
    \State $\pi^{t+1} = \argmin_{\pi \in \Pi} \sum_{\tau=1}^t u(\pi, g^t) + \phi(\pi)$. \algcommentlight{Regularized cost-sensitive oracle.}
    \EndFor
    \Return $\pi^{t^\ast}$ with $t^\ast = \argmin_{t \in [T]} u^t.$
    \end{algorithmic}
    \label{alg:fail_inter}
\end{algorithm}

Now we first show the guarantee on the no-regret procedure. Note that this result does not immediately imply the return policy is close to the offline distribution, because our data collection distribution is: we first roll in our policy, and then switch to the offline distribution for one-step. The no-regret guarantee is that the policy distribution will be close to this one-step-shift distribution. In the following, whenever we refer to the admissibility assumption \pref{assum:admissibility_ipm} or bellman completeness assumption \pref{assum:completeness}, we refer to their stationary version analogue.

\begin{lemma}[Guarantee of the no-regret procedure] \label{lem:noregret}
Assume \pref{assum:admissibility_ipm} holds. Suppose that $\Gcal$ is the discriminator class, $\rho$ is the roll in distribution, then let $\pif$ be the return policy from running \pref{alg:fail_inter} for $T$ iterations, we have with probability at least $1-\delta$,
\begin{align*}
    d_{\Gcal_{h}}(\pi \mid \pi, \pi \circ \mu) \leq \min_{\pi' \in \Pi}d_{\Gcal_{h}}(\pi' \mid \pi', \pi' \circ \mu) + \epsforward(\delta, T),
\end{align*} 
where 
\begin{align*}
    \epsforward(\delta, T) = 8\sqrt{\frac{2A^2\log(2|\Gcal_{h}||\Pi|/\delta)}{T}}.
\end{align*}
\end{lemma}
The proof uses the same concentration argument used in the proof of \pref{lem:concentration} and the same no-regret techniques that handle non-stationary distributions in \citet{vemula2023virtues} so we omit the proof here. 

Then we can use the no-regret guarantee for the final result for the forward algorithm:
\begin{theorem}[Guarantee of \pref{alg:fail_inter}]\label{thm:fail_inter}
Assume \pref{assum:completeness} and \pref{lem:noregret} hold with probability at least $1-\delta$. Then the returned policy $\pif$ of \pref{alg:fail_inter} after $T$ iterations satisfies that,
\begin{align*}
    \IPM_{\Gcal}(\pif, \mu) \leq \frac{\epsforward(\delta, T) + \varepsilon^{\mathsf{be}}}{1-\gamma}.
\end{align*}
\end{theorem}

\begin{proof}
We start with an important identity: for any stationary policy $\pi$, we have 
\begin{align*}
    \EE_{s \sim d^{\pi}} [f(s)] = (1-\gamma) \EE_{s \sim P_0} [f(s)] + \gamma \EE_{s \sim d^\pi, a \sim \mu, s' \sim P(s,a)}[f(s')].
\end{align*}
Then we have 
\begin{align*}
    \IPM_{\Gcal}(\pif, \mu) =& \max_{g \in \Gcal} \abr{\EE_{s \sim d^{\pif}}[g(s)] - \EE_{s \sim \mu}[g(s)]} \\
    =& \max_{g \in \Gcal} \abr{\gamma \EE_{s \sim d^{\pif}, a \sim \pif(s), s' \sim P(s,a)}[g(s')] - \gamma \EE_{s \sim \mu, a \sim \mu(s), s' \sim P(s,a)}[g(s')]} \\
    \leq& \max_{g \in \Gcal} \abr{\gamma \EE_{s \sim d^{\pif}, a \sim \pif(s), s' \sim P(s,a)}[g(s')] - \gamma \EE_{s \sim d^{\pif}, a \sim \mu(s), s' \sim P(s,a)}[g(s')]}  + \\ &
    \max_{g \in \Gcal} \abr{\gamma \EE_{s \sim d^{\pif}, a \sim \mu(s), s' \sim P(s,a)}[g(s')] - \gamma \EE_{s \sim \mu, a \sim \mu(s), s' \sim P(s,a)}[g(s')]}.
\end{align*}
Note that the first term, by the no-regret guarantee in \pref{lem:noregret}, is bounded by $\epsforward$, and the second term we can bound by the following, which is similar to the technique we use in the proof for the forward run in the non-stationary setting:

Now denote 
\begin{align*}
    g^\ast := \argmax_{g \in \Gcal} \abr{\gamma \EE_{s \sim d^{\pif}, a \sim \mu(s), s' \sim P(s,a)}[g(s')] - \gamma \EE_{s \sim \mu, a \sim \mu(s), s' \sim P(s,a)}[g(s')]}
\end{align*}
we let 
\begin{align*}
    g^{\mathsf{b}} = \argmin_{g \in \Gcal} \nbr{g - \Tcal g^\ast}_\infty,
\end{align*}
the bellman backup of $g^\ast$ under the offline distribution backup, then we have 
\begin{align*}
    &\max_{g \in \Gcal} \abr{\gamma \EE_{s \sim d^{\pif}, a \sim \mu(s), s' \sim P(s,a)}[g(s')] - \gamma \EE_{s \sim \mu, a \sim \mu(s), s' \sim P(s,a)}[g(s')]}\\
    =& \gamma\abr{ \EE_{s \sim d^{\pif}, a \sim \mu(s), s' \sim P(s,a)}[g^\ast(s')] -  \EE_{s \sim \mu, a \sim \mu(s), s' \sim P(s,a)}[g^\ast(s')]} \\
    \leq& \gamma\abr{ \EE_{s \sim d^{\pif}}[g^\ast(s)] -  \EE_{s \sim \mu}[g^\ast(s)]} + \varepsilon^{\mathsf{be}}\\
    =& \gamma \IPM_{\Gcal}(\pif, \mu) + \varepsilon^{\mathsf{be}}.
\end{align*}
Finally, putting everything together we will get:
\begin{align*}
    \IPM_{\Gcal}(\pif, \mu) \leq \epsforward + \gamma \IPM_{\Gcal}(\pif, \mu) + \varepsilon^{\mathsf{be}},
\end{align*}
which by rearranging we get:
\begin{align*}
    \IPM_{\Gcal}(\pif, \mu) \leq \frac{1}{1-\gamma} \epsforward + \varepsilon^{\mathsf{be}}.
\end{align*}
\end{proof}

Finally, to obtain the result for the stationary version, we can simply combine the result of \pref{thm:cpi_trace}
and \pref{thm:fail_inter}. We obtain the stationary analog of \pref{thm:foobar} by replacing the horizon dependency with the effective horizon $\frac{1}{1-\gamma}$. We remark that, in the stationary setting, the name ``forward phase'' and ``backward phase'' may not be as clear as in the non-stationary setting, but one can interpret the ``forward phase'' as the offline distribution matching phase, and the ``backward phase'' as the policy refinement (optimization) phase. 

\clearpage
\section{Discussion on the Structural Assumption}\label{app:structural}
Here we give the formal introduction of the structural assumption. We adopt the one from \citet{du2021bilinear} as it is the structural assumption made in the most hybrid RL analysis \citep{song2022hybrid,nakamoto2023cal}. However, the results will transfer trivially to similar structural assumptions like Bellman Eluder dimension \citep{jin2021bellman} or coverability \citep{xie2023the}. In other hybrid RL works, \citet{wagenmaker2023leveraging} assumes linear MDPs structure \citep{jin2020provably} and \citet{li2023reward} assumes tabular MDPs. 

\begin{definition}[{Bilinear model \citep{du2021bilinear}}] 
\label{def:bilinear_model}
We say that the MDP together with the function class $\Fcal$ is a bilinear model of rank \(d\) if for any \(h \in [H-1]\), there exist two (unknown) mappings $X_h,W_h:\Fcal\mapsto \mathbb{R}^d$ with $\max_{f}\| X_h(f)\|_2 \leq B_X$ and $\max_{f} \|W_h(f)\|_2 \leq B_W$ such that:
\begin{align*} 
\forall f, g\in \Fcal: \; \abr{\EE_{s, a \sim d_h^{\pi^f}}\sbr{  g_h(s,a) - \Tcal g_{h+1}(s,a) } } = \abr{ \inner{ X_h(f)}{ W_h(g) } }.
\end{align*} 
\end{definition}

Note that the dimension of the mapping $X$ and $W$ are called the bilinear rank, which is bounded by $d$. For example, in tabular MDPs, $d = SA$, and in linear MDPs \citep{jin2020provably} and low-rank MDPs \citep{agarwal2020flambe}, $d$ is the dimension of the feature vector. 

Continuing from \pref{remark:structural}, suppose we are
in the tabular setting, since we involve function approximation, the worst-case log size of the function class will still be bounded by $SA$, and then the final bound will be worse than the tightest bound in the tabular case \citep{zhang2023settling}. Note that in the worst case, the $S$ dependency is unavoidable in the hybrid RL setting even with canonical offline data (see Theorem 3 of \citet{xie2021policy}). However, we argue that the dependency of $SA$ has a different source compared to the tightest analysis in tabular MDPs such as \citet{azar2017minimax,zhang2023settling}: the $SA$ dependency in these analyses is from the fundamental complexity measure $d=SA$ in the MDP itself. For example, in the worst case, one has to hit each state-action pairs enough times such that the confidence intervals shrink. On the other hand, the size of the function class is not necessarily tied to the complexity of dynamics or rewards, and the $SA$ dependency of the log size of the function class is always avoidable with the right choice of function class (inductive bias). However, unlike our analysis, such $SA$ dependency still shows up in the current hybrid RL analysis, where their suboptimality scales in (ignoring irrelevant terms):
\begin{align*}
    V^{\picomp} - V^{\pi} \leq O\rbr{\ccov(\picomp) \sqrt{\frac{d \log(|\Fcal|/\delta)}{N}}} = O \rbr{\ccov(\picomp) \sqrt{\frac{SA \log(|\Fcal|/\delta)}{N}}},
\end{align*}
i.e., previous results pay for both $SA$ and $\log(|\Fcal|)$.

Our result is even more favorable in the more general cases. In the main text, we use the $\ell_\infty$ coverage for the simplicity of presentation, which may be unbounded when the state space is not finite. Here we introduce a tighter coverage coefficient that is similar to the previous \emph{expected} Bellman error coverage used in offline RL \citep{xie2021bellman} and hybrid RL \citep{song2022hybrid,nakamoto2023cal}, which we called \emph{performance difference coverage}:
\begin{definition}[Performance difference coverage]
    For the given offline distribution $\rho$, and for any policy $\pi$, the performance difference coverage coefficient is define as 
    \begin{align*}
    \ccovpd(\pi) = \max_{\pi' \in \Pi^{\mathsf{det}}} \frac{\sum_{h=1}^H \EE_{s_h \sim d^\pi_h} \sbr{\max_a A^{\pi'}_h(s_h,a)}}{\sum_{h=1}^H \EE_{s_h \sim \mu_h} \sbr{\max_a A^{\pi'}_h(s_h,a)}}.    
    \end{align*}
\end{definition}

With this we can state the following more refined result:
\begin{theorem}\label{thm:foobar_cov}
    Suppose \pref{assum:admissibility_ipm,assum:completeness,assum:real} hold. Then with probability $1-\delta$, the returned policy $\pib_{1:H}$ from \pref{alg:foobar} with discriminator constructed from \pref{eq:discriminator}, $N^{\mathsf{for}}$ offline and forward samples, and $N^{\mathsf{back}}$ backward samples, satisfies that for any comparator policy $\picomp$ such that $\ccovpd(\picomp) < \infty$,
    \begin{align*}
        V^{\picomp} - V^{\pib} \leq \varepsilon,
    \end{align*}
    when 
    \begin{align*}
        N^{\mathsf{for}} = O\rbr{\frac{{\ccovpd}^2(\picomp)H^4A\log(|\Fcal||\Pi|/\delta)}{\varepsilon^2}},~~ N^{\mathsf{back}} = O\rbr{\frac{{\ccovpd}^2(\picomp)H^4A\log(|\Fcal|/\delta)}{\varepsilon^2}}.
    \end{align*}
\end{theorem}
The proof is the same as the proof of \pref{thm:foobar}, and one can check we can safely replace $\ccov$ with $\ccovpd$ during the distribution shift step. Note that this result does not depend on specific structural complexity measures of the MDPs (e.g., the bilinear rank \citep{du2021bilinear,song2022hybrid}). On the other hand, one advantage of previous hybrid RL algorithms is that they work under situations where the offline data is inadmissible (c.r. \pref{tab:comparison}). 

Intuitively, the bilinear rank assumption captures the following idea: the rank $d$ denotes the number of ``distribution shift'' that the algorithm will encounter during the online policy or value function update, i.e., how many times the algorithms have to roll out so that the previous data distribution will cover the current policy's visitation distribution. However, in \Foobar{}, there is no distribution issue (because for every horizon, we will collect some data, train a one-step policy, commit to the policy, and not update it anymore). We believe the absence of the distribution shift problem is partially due to the admissibility assumption we make for the offline dataset, but an understanding of the fundamental connections between the admissibility and structural assumptions remains an interesting open problem.

Finally, we remark that there is one previous hybrid RL work that is also free from the structural assumption, which is \citet{zhou2023offline}. However, like the previous line of works that study RL in the reset model \citep{kakade2002approximately,bagnell2003policy}, their analysis requires an exploratory reset distribution, which is as strong as having a reset model.

\clearpage
\section{Experiment Details}\label{app:exp} 
\begin{figure}[th]
    \centering
    \includegraphics[width=0.3\textwidth]{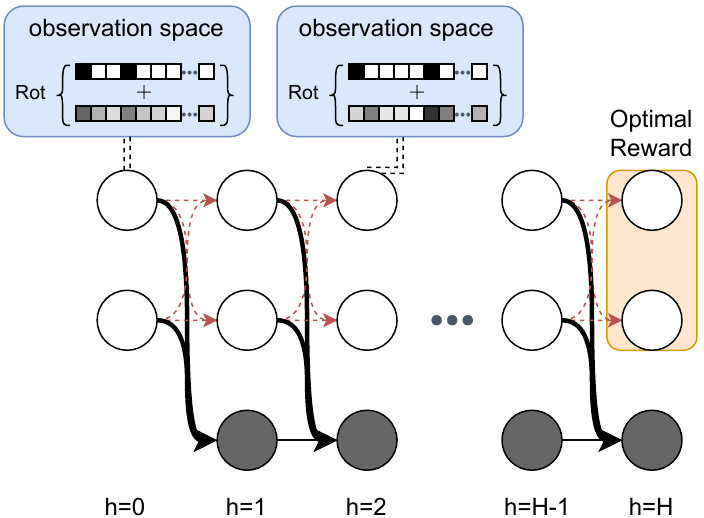}
    \hspace{2cm}
    \includegraphics[width=0.22\textwidth]{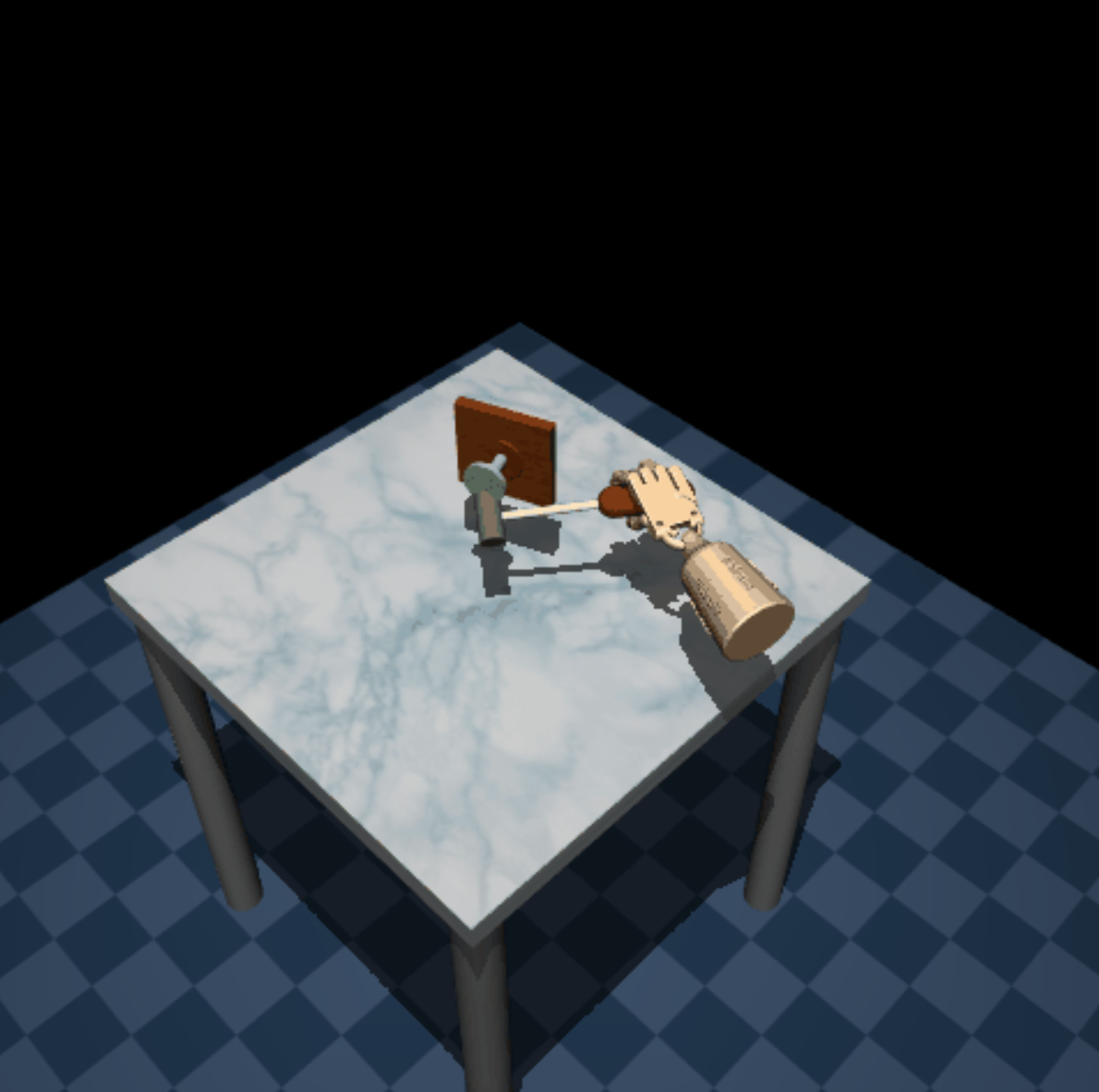}
    \caption{Visualization of the environment. Left: combination lock. Right: hammer. The left figure is reproduced from \citet{zhang2022efficient} with permission from the authors.} 
    \label{fig:envs}
\end{figure}

\subsection{Combination Lock Environment}\label{app:lock_intro}

\paragraph{Detailed environment description} In our experiment, the \textit{diabolical combination lock} problem served as the testing ground for our algorithm. This scenario is defined by a horizon of length $H$ and involves a selection from 10 distinct actions. At each point in the sequence, denoted as step $h$, the system can be in one of three potential hidden states, symbolized as $z_{i;h}$ for $i$ values in the set ${0,1,2}$. States $z_{i;h}$, where $i$ falls within ${0,1}$, are considered advantageous, while the state $z_{2;h}$ is categorized as disadvantageous.

For each advantageous state $z_{i;h}$ (where $i$ is either 0 or 1), an action, denoted as $a_{i;h}$, is chosen randomly from the pool of 10 actions. In such states, executing the action $a_{i;h}$ leads to a transition to either state $z_{0;h+1}$ or $z_{1;h+1}$, with each possibility having an equal chance of occurrence. Choosing any action other than $a^\ast_{i;h}$ in these states ensures a move to the state $z_{2;h+1}$. In the state $z_{2;h}$, the agent's action choice does not affect its transition, which is always to $z_{2;h+1}$.

The reward structure is such that a reward of 1 is assigned at state $z_{i;H}$ for $i \in {0,1}$. There is also a 50\% probability of receiving a minor, inverted reward of 0.1 when transitioning from a favorable to an unfavorable state. All other state transitions or states do not yield any reward.

Observations in this problem, denoted as $s$, have a dimensionality of $2^{\lceil{\log(H+4)}\rceil}$. This is formulated by concatenating the one-hot vectors representing the hidden state $z$ and the horizon $h$, to which noise from the distribution $\Ncal(0,0.1)$ is added for each dimension. This is then adjusted with zeroes where necessary and processed through a Hadamard matrix. The starting state distribution is uniformly divided among $z_{i;0}$ for $i \in {0,1}$. An important aspect to note is that the ideal strategy involves consistently selecting the action $a^*_{i;h}$ at each step $h$. Once the agent enters a disadvantageous state, it remains in such state till the episode concludes, thus forfeiting the opportunity for a significant end reward. This presents a significant challenge in terms of exploration, as a strategy based on random uniform selection yields only a $10^{-H}$ chance of reaching the intended goals.

\begin{figure}[th]
    \centering
    \includegraphics[width=0.5\textwidth]{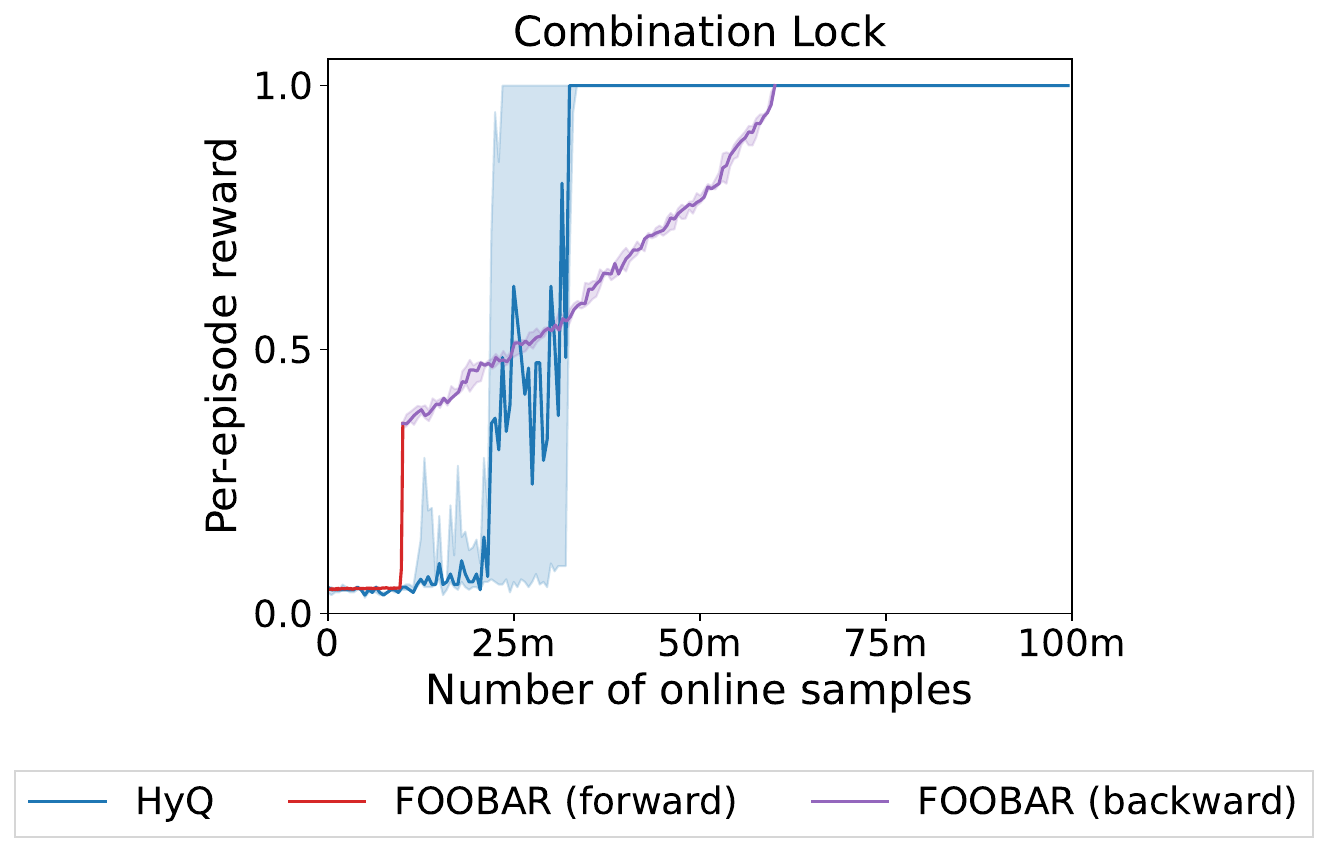}
    \caption{Zoomed-in training curve of \Foobar{}.}
    \label{fig:zoom}
\end{figure}

\paragraph{Implementation details} We parametrize the forward policy with a 2-layer neural network with Tanh activation and we model the action distribution with diagonal Gaussian. For the backward policy, we use least square regression to estimate the Q-functions where we follow the same parametrization as in \citet{song2022hybrid}. We use the median trick \citep{fukumizu2009kernel} to set up the bandwidth for the RBF kernel. Hyperparameters for the combination lock experiment are presented in \pref{table:lock}.

\begin{table}[htp]
  \caption{Hyperparameters for combination lock}
  \centering
  \begin{tabular}{cc}
    \toprule
    &\; Value  \\
    \hline
    Offline sample size (per horizon)           &\; 2000     \\
    Online forward sample size (per horizon)    &\; 2000     \\
    Forward policy hidden layer size            &\; 128      \\
    Min-max game iteration                      &\; 1000     \\
    Online backward sample size (per horizon)   &\; 5000     \\
    Backward number of gradient descent updates &\; 1500     \\
    Backward minibatch size                     &\; 128      \\
    Learning rate                               &\; 0.001    \\
    \toprule
\end{tabular}\label{table:lock}\end{table}

For completeness, here we also provide a zoomed-in training curve for \Foobar{} with both forward phase and backward phase labeled in \pref{fig:zoom}.

\begin{table}[htp]
  \caption{Hyperparameters for hammer}
  \centering
  \begin{tabular}{cc}
    \toprule
    &\; Value  \\
    \hline
    Offline sample size (per horizon)           &\; 2000     \\
    Online forward sample size (per horizon)    &\; 2000     \\
    Forward policy hidden layer size            &\; 128      \\
    Min-max game iteration                      &\; 1000     \\
    Online backward sample size (per horizon)   &\; 5000     \\
    Backward number of gradient descent updates &\; 1500     \\
    Backward minibatch size                     &\; 128      \\
    Learning rate                               &\; 0.001    \\
    \toprule
\end{tabular}\label{table:hammer}\end{table}

\subsection{Hammer}\label{app:hammer}

For the offline dataset construction of the hammer environment, we use the expert offline dataset provided in the d4rl benchmark. Take the first 2000 trajectories and extract the first 50 horizons for each trajectory for the offline dataset. Note that for the hammer environment, the expert dataset does not contain the optimal policy, and in fact only 80\% trajectories of offline datasets contain a successful state at horizon 50. We use the expert offline dataset mainly due to the fact that this dataset is collected by a diagonal Gaussian policy, which is the same as our parametrization of the policy so admissibility holds gracefully. However, we believe using the recently proposed diffusion policy \citep{block2023provable} will address this issue since diffusion policies can parameterize multimodal distributions. 

As mentioned in the main text, we make one modification on the forward phase that for each horizon, we iterate between optimizing the forward policy and using the latest policy to collect more data. During training, we still use the same min-max objective and instead of performing importance weighting on the uniform policy, we adjust the importance weight with respect to the data collection policy. For the backward run, we follow \pref{alg:cpi_trace}: we roll-in the forward policy to a random horizon, and we switch to the current stationary SAC policy to roll out and only update the SAC policy using the data collected during the roll out. We provide the hyperparameter table in \pref{table:hammer}.

\begin{figure}
    \centering
    \includegraphics[width=0.18\textwidth]{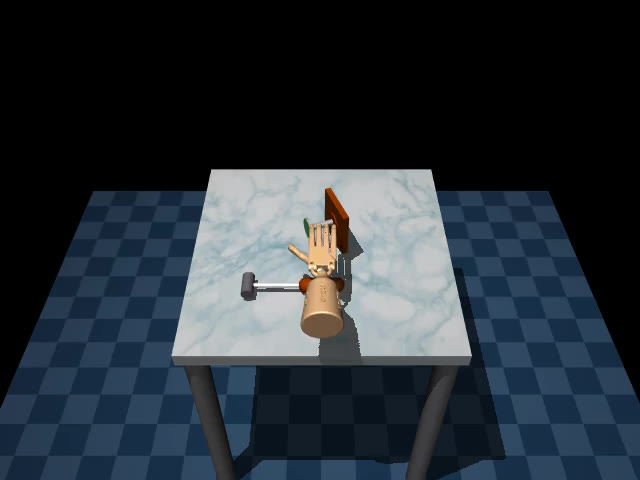}
    \includegraphics[width=0.18\textwidth]{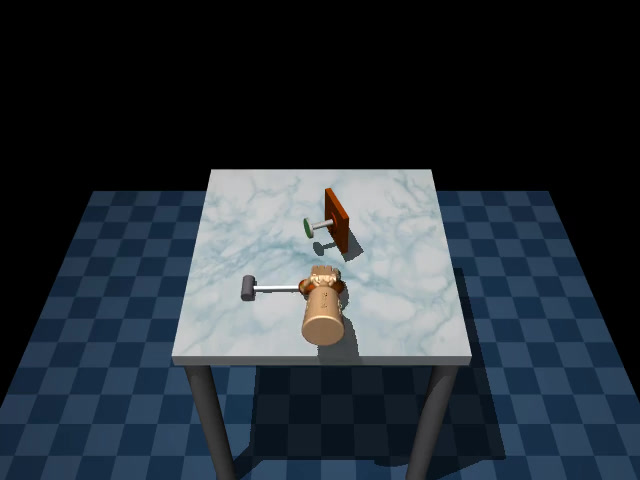}
    \includegraphics[width=0.18\textwidth]{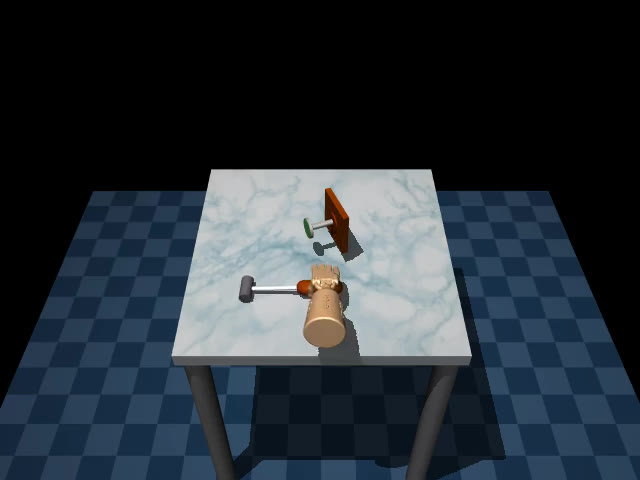}
    \includegraphics[width=0.18\textwidth]{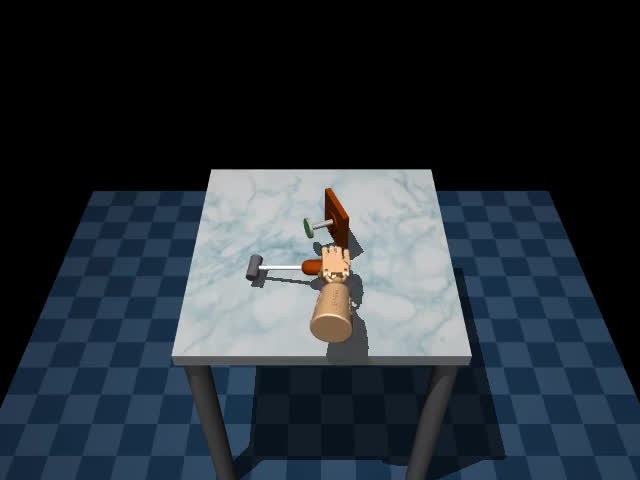}
    \includegraphics[width=0.18\textwidth]{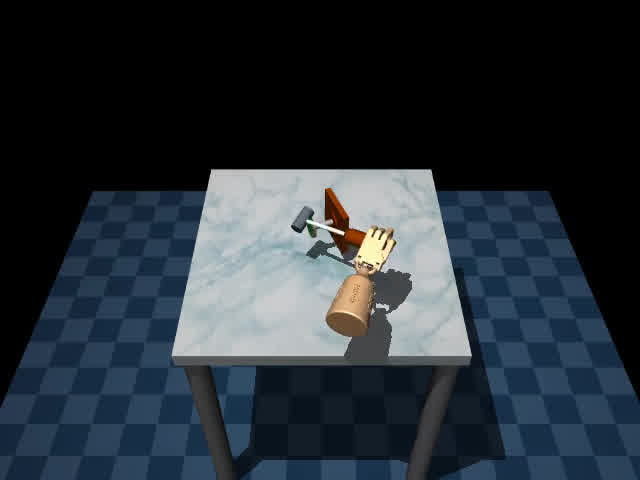}
    \caption{Visualization of a typical failure trajectory of the forward policy. Note that the hammer hits the nail but does not fully push the nail into the board.}
    \label{fig:hammers}
\end{figure}

\begin{figure}
    \centering
    \includegraphics[width=0.4\textwidth]{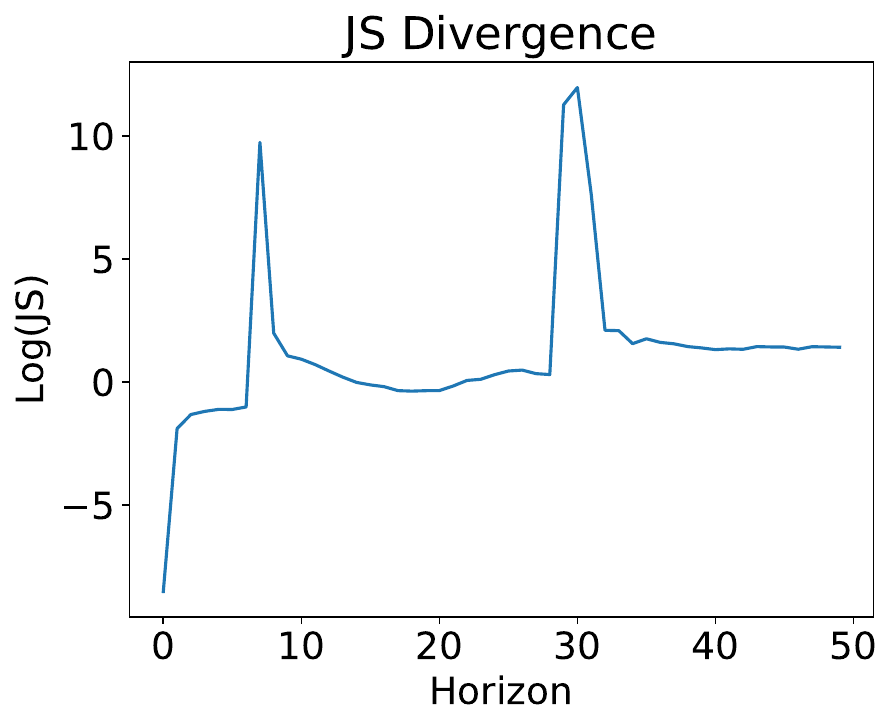}
    \caption{Plot of empirical JS divergence between forward policy and offline data for each horizon. The $y$-axis is in the natural log scale.}
    \label{fig:js}
\end{figure}

To show the performance of the forward run, we notice that on average the forward policies will have $10\%$ success rate at the end of the forward phase (compared to the $80\%$ success rate in the offline dataset). However, it is due to the strict success evaluation of the hammer-binary environment, and we note that even if the policy fails to solve the task, it still covers the optimal policy reasonably, and thus although in theory, the IPM between forward policy and offline distribution may not be small, the forward policy still covers the optimal policy, and the learning will success due to \pref{thm:foobar_cov}. Here we give a qualitative and quantitative evaluation of the forward policy. For the qualitative evaluation, we visualize a typical failure trajectory of the forward policy in \pref{fig:hammers} and note that the hammer hits the nail but does not fully push the nail into the board. For the qualitative evaluation, we test the empirical Jensen-Shannon (JS) divergence between the dataset induced by the forward policy and the offline dataset, and we plot the average across the 10 runs in natural log scale in \pref{fig:js}.

\subsection{Inadmissble Offline}
In this section, we describe the construction of the experiments in \pref{sec:exp_inadm}. For the benign inadmissibility setting, we collect the offline data in the following way: we reset the initial state distribution the same way as regular combination lock, and for horizon $h=1$, we generate the observation of state 0 (good state), state 1 (good state) and state 2 (bad state) with probability $(0.1, 0.05, 0.85)$ respectively. For $h\geq 2$, we generate the observation of state 0 (good state), state 1 (good state) and state 2 (bad state) with probability $(0.5 \cdot h,\;0.05 \cdot h,\;1 - 0.1 \cdot h)$ respectively. Note that this is an inadmissible offline dataset because the probability of visiting good states is non-increasing over the horizon for any admissible distribution. 

For the adversarial inadmissibility setting, the offline distribution follows the same construction as the benign setting: we reset the initial state distribution the same way as regular combination lock, and for timestep $h=1$, we generate the observation of state 0 (good state), state 1 (good state) and state 2 (bad state) with probability $(0.1, 0.05, 0.85)$ respectively. For $h\geq 2$, we generate the observation of state 0 (good state), state 1 (good state) and state 2 (bad state) with probability $(0.5 \cdot h,\;0.05 \cdot h,\;1 - 0.1 \cdot h)$ respectively. However, we modify $P_1$ and $P_2$ of the combination lock in the following way: at timestep 1, taking good actions in either state 0 or state 1 will have a 0.1 probability transiting to state 0 in timestep 2, and 0.9 probability to state 1; taking any bad action will have a probability of 0.05 transiting to state 1, and 0.85 probability to transit to state 2. However, in timestep 2, only state 0 will be treated as a good state, and state 1 will be treated as a bad state and thus taking any action in state 1 in timestep 2 will transit to state 2 deterministically. All the remaining dynamics are the same as the regular combination lock. We note that this is exactly the same construction as in \pref{prop:hard_tv}, and the optimal policy will have a success rate of $10\%$ due to the stochasticity of the environment. 

Finally, we include the hyperparameters for each baseline in \pref{table:foobar_inadm} and \pref{table:psdp_inadm}.

\begin{table}[htp]
  \caption{Hyperparameters for \Foobar{}}
  \centering
  \begin{tabular}{cc}
    \toprule
    &\; Value  \\
    \hline
    Offline sample size (per horizon)           &\; 2000     \\
    Online forward sample size (per horizon)    &\; 2000     \\
    Forward policy hidden layer size            &\; 128      \\
    Min-max game iteration                      &\; 1000     \\
    Online backward sample size (per horizon)   &\; 4000     \\
    Backward number of gradient descent updates &\; 2000     \\
    Backward minibatch size                     &\; 128      \\
    Learning rate                               &\; 0.001    \\
    \toprule
\end{tabular}\label{table:foobar_inadm}\end{table}

\begin{table}[htp]
  \caption{Hyperparameters for \Psdp{}}
  \centering
  \begin{tabular}{cc}
    \toprule
    &\; Value  \\
    \hline
    Offline sample size (per horizon)           &\; 2000     \\
    Online  sample size (per horizon)           &\; 4000     \\
    Number of gradient descent updates          &\; 2000     \\
    Minibatch size                              &\; 128      \\
    Learning rate                               &\; 0.001    \\
    \toprule
\end{tabular}\label{table:psdp_inadm}\end{table}

\end{document}